\documentclass[pdflatex,sn-mathphys-num]{sn-jnl}

\usepackage{graphicx}%
\usepackage{subcaption}
\usepackage{multirow}%
\usepackage{amsmath,amssymb,amsfonts}%
\usepackage{amsthm}%
\usepackage{mathrsfs}%
\usepackage[title]{appendix}%
\usepackage{xcolor}%
\usepackage{textcomp}%
\usepackage{manyfoot}%
\usepackage{booktabs}%
\usepackage{algorithm}%
\usepackage{algpseudocode}%
\usepackage{algorithmicx}%
\usepackage{listings}%
\usepackage{threeparttable}
\usepackage{tabularx}
\usepackage{rotating}

\theoremstyle{thmstyleone}%
\theoremstyle{thmstyletwo}%

\theoremstyle{thmstylethree}%

\begin{document}

\title[Article Title]{Anatomy-preserving unpaired cone-beam CT refinement for image-guided radiotherapy using pseudo-label guided diffusion}

\author[1]{\sur{Qi Lai}}\email{q.lai@siat.ac.cn}

\author[2]{\sur{YuTong He}}\email{heyt3@mail.sustech.edu.cn}




\affil[1]{\orgdiv{Shenzhen Institutes of Advanced Technology}, \orgname{Chinese Academy of Sciences}, \orgaddress{ \city{Shenzhen}, \postcode{518055},\country{China}}}

\affil[2]{\orgdiv{Southern university of science and technology}, \orgaddress{\city{Shenzhen}, \postcode{518055},\country{China}}}



\affil[3]{%
  \orgdiv{Department of Neurosurgery},
  \orgname{Huashan Hospital, Fudan University},
  \orgaddress{
    \city{Shanghai},
    \postcode{200040},
    \country{China}}}


 \abstract{
Cone-beam computed tomography (CBCT) is widely used in image-guided radiotherapy, yet its broader clinical utility is limited by scatter, beam hardening, noise, truncation, and other artifacts that degrade image quality and CT-number fidelity. Although deep learning methods have shown promise for CBCT correction, most rely on paired CBCT–CT training data, which are difficult to obtain in routine clinical practice because of motion, anatomical variation, and acquisition mismatch. Here we present RefineCBCT, an unpaired CBCT refinement framework that combines pseudo-label guidance with short-step diffusion to suppress artifacts while preserving patient-specific anatomy. By formulating CBCT correction as anatomy-preserving refinement rather than unconstrained image translation, RefineCBCT provides structurally consistent guidance for diffusion-based restoration under unpaired supervision.

RefineCBCT was trained and evaluated using unpaired CBCT and planning CT data from the public LUNG TCIA and PELVIC TCIA datasets and compared with representative GAN-based and diffusion-based methods. On LUNG TCIA, RefineCBCT achieved the best performance across all evaluated metrics, with a mean absolute error of 19.411, root mean squared error of 62.758, peak signal-to-noise ratio of 30.845 dB, and structural similarity index of 0.931. On PELVIC TCIA, it achieved the best mean absolute error, peak signal-to-noise ratio, and structural similarity index, with values of 14.905, 36.671 dB, and 0.876, respectively. The refined images showed reduced streaking and shading artifacts, clearer anatomical boundaries, and improved soft-tissue uniformity. Line-profile and region-of-interest analyses further demonstrated closer agreement with planning CT in both bone and soft-tissue regions.

These findings suggest that RefineCBCT enables effective and computationally efficient CBCT refinement under clinically realistic unpaired training conditions, and may support more reliable use of CBCT in image-guided radiotherapy workflows. Code is publicly available at \url{https://github.com/Lai-KK/RefineCBCT.git}, and the evaluated datasets are publicly available from The Cancer Imaging Archive.
}




\maketitle

\section{Introduction}
\label{sec:Introduction}

Cone-beam computed tomography (CBCT) is routinely used in image-guided radiotherapy because it provides fast, on-board volumetric imaging during treatment \cite{zhao2026generative,kapila2011current,purdie2007cone,hvid2018cone,allen2023evaluating}. However, despite its convenience and widespread availability, CBCT is still used mainly for patient setup and coarse localization rather than for quantitative or decision-critical applications \cite{zhang2023breath,zhong2023bi,wu2022two,weiss2019cone,quereshy2008applications,de2009cone,hirota2006cone}. 
The main reason is that CBCT suffers from severe image degradation, including scatter, beam hardening, noise, truncation, and motion-related corruption, which lead to streaking artifacts, shading, blurred boundaries, low soft-tissue contrast, and inaccurate intensities  \cite{hsieh2000iterative,jarry2006characterization,zhang2007reducing,sisniega2017motion,song2025ultra}. 
These limitations reduce confidence in downstream tasks such as contouring, dose evaluation, adaptive planning, and automated analysis. 
Therefore, the key clinical question is not whether CBCT can be acquired, but whether routinely acquired CBCT can be refined into an image that is reliable enough for broader clinical use \cite{lin2023novel,su2023super}.

A straightforward solution is to learn a mapping from low-quality CBCT to CT-like images \cite{jin2024staran,chao2023psdp}. 
However, most existing learning-based methods rely on paired CBCT-CT supervision \cite{wu2026unsupervised}, which is difficult to obtain in real clinical settings \cite{zhang2024review}. CBCT and CT are often acquired at different time points and under different conditions, with anatomical discrepancies caused by posture changes, organ motion, or disease progression \cite{yang2025deep}. Even after deformable registration, residual mismatch can introduce incorrect supervision and compromise structural fidelity during training \cite{chen2026neural}. As a result, paired learning is not only data-expensive but also fundamentally limited by imperfect correspondence.

This makes \textit{unpaired} CBCT refinement more clinically realistic, but also substantially more challenging. Without voxel-wise targets, the model must suppress artifacts while preserving patient-specific anatomy. In medical imaging, this requirement is crucial: a visually cleaner image is not useful if clinically relevant structures are hallucinated, displaced, or over-smoothed. Existing unpaired translation methods can improve appearance, but often lack sufficient control over anatomical consistency and restoration stability \cite{zhu2017unpaired,bashkirova2019adversarial,you2020bayesian}. Diffusion models offer strong generative priors and powerful restoration capability, but in unpaired CBCT enhancement they still need effective guidance to avoid drifting away from the input anatomy, and their iterative sampling cost can limit practical deployment \cite{syndiff,FGDM,chung2023solving,zhao2024wia,li2025tlir}.

\begin{figure}[!ht]
    \centering
    \includegraphics[width=\linewidth]{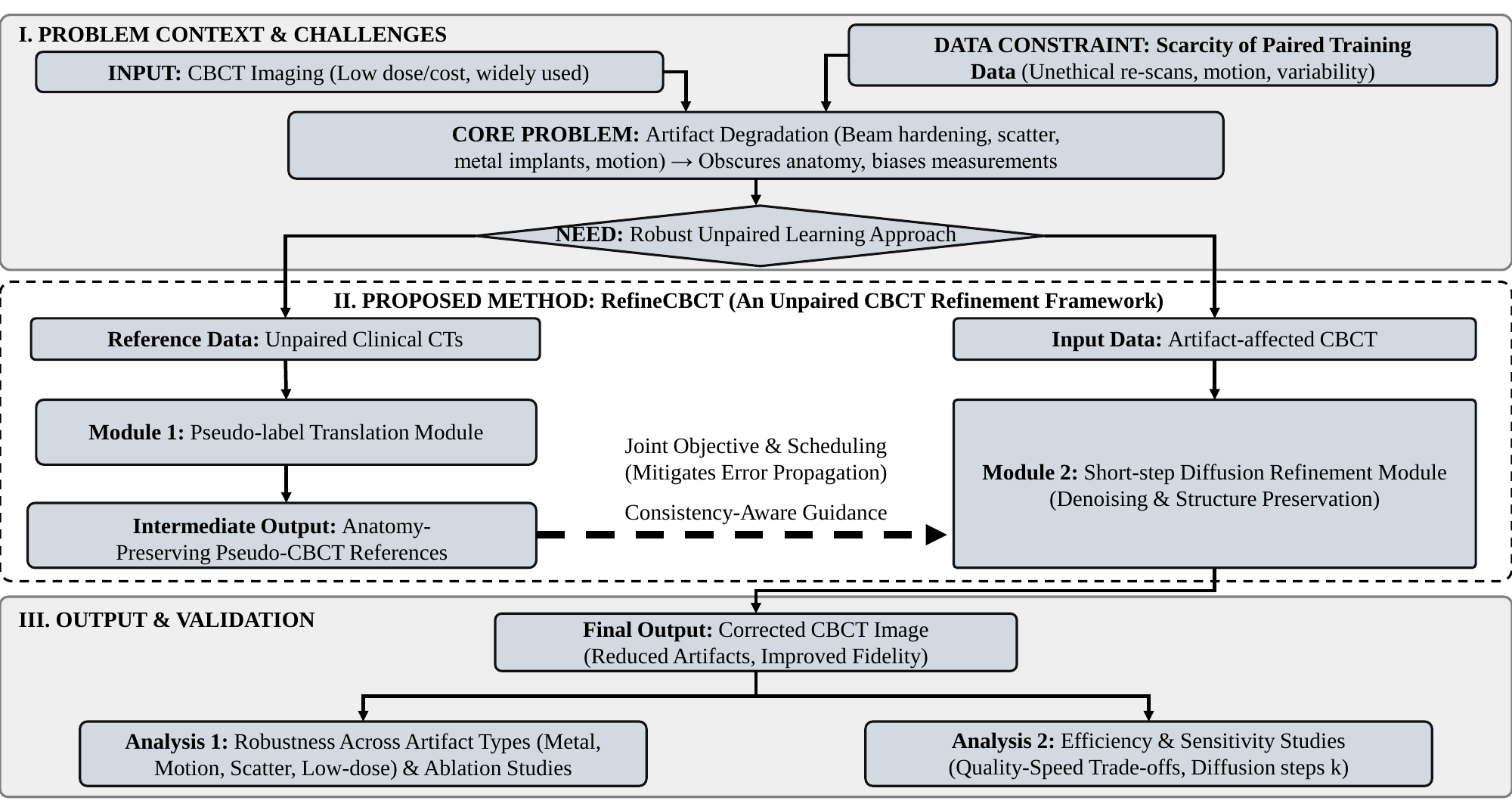}
    \caption{Clinical motivation and conceptual overview of RefineCBCT. Artifact-degraded CBCT and the scarcity of accurately paired training data motivate an unpaired refinement strategy. RefineCBCT couples a cycle-consistent pseudo-label translation module with a short-step diffusion refinement module. The translated reference provides anatomy-aware guidance, while short-step diffusion suppresses artifacts and preserves patient-specific structures with reduced sampling cost.}
    \label{fig:framework_overview}
\end{figure}

To address these issues, we propose \textbf{RefineCBCT}, an unpaired CBCT refinement framework guided by pseudo labels and short-step diffusion, as shown in Fig.~\ref{fig:framework_overview}. The method is built on a simple idea: \textit{unpaired} refinement should be constrained enough to preserve anatomy, but flexible enough to remove complex artifacts. To achieve this, RefineCBCT first uses a \textit{pseudo-label translation module} to generate anatomy-preserving intermediate references from unpaired clinical CT data. These pseudo labels provide a stable supervisory bridge in the absence of paired CBCT-CT correspondence. RefineCBCT then applies a \textit{short-step diffusion refinement module} to perform controlled denoising and artifact suppression, improving image fidelity while avoiding unconstrained image generation. This design makes the framework not only anatomically aware, but also more computationally efficient than long-horizon diffusion restoration, which is important for clinical usability.

The methodological highlight of RefineCBCT lies in the explicit coupling of designing around three priorities that are often difficult to satisfy simultaneously in unpaired medical image enhancement: \textit{fidelity}, \textit{robustness}, and \textit{efficiency}.
The translated pseudo-label is not treated as a final synthetic image or as fixed offline supervision. Instead, it provides a structural constraint that is jointly updated with the diffusion model. The reconstruction objective preserves CBCT intensity information, the cycle-consistency objective maintains cross-domain anatomy, and the time-conditional adversarial objective regularizes short-step diffusion transitions. This joint design is intended to prevent errors in the pseudo-label from propagating directly to the corrected image, thereby balancing artifact suppression, anatomical fidelity, and computational efficiency.
The main contributions are as follows:

\noindent i) We propose \textbf{RefineCBCT}, a CBCT enhancement framework that formulates the task as an anatomically constrained refinement problem and learns from unpaired CT and CBCT data \textit{without} requiring voxel-wise aligned supervision. Specifically, the method integrates a consistency-aware pseudo-label guidance strategy, which couples bidirectional CT--CBCT translation with conditional diffusion to guide artifact suppression while reducing structural distortion and hallucination, together with a short-step diffusion refinement process that lowers sampling cost while maintaining stable restoration. In this way, RefineCBCT achieves anatomically faithful and computationally efficient CBCT refinement, offering a practical balance between image quality, structural preservation, and clinical usability.

\noindent ii) We validate the proposed approach through comprehensive experiments across thoracic and pelvic CBCT datasets, demonstrating improved quantitative performance, HU fidelity, and visual quality, as well as favorable runtime efficiency and downstream utility. These results support the potential clinical value of RefineCBCT and indicate that it provides an effective and clinically relevant solution for CBCT enhancement beyond the limitations of paired-data supervision.

\section{Results}
\label{sec2}
\subsection{Experimental Setup} \label{ssec:setup}

Experiments were conducted on two public datasets from The Cancer Imaging Archive (TCIA) \cite{r13,r14,r15,r16}, each containing cone-beam CT (CBCT) and planning CT (pCT) scans acquired from the same patient cohort. Although CBCT and pCT data were available within each dataset, the proposed method was trained under an unpaired setting to reflect realistic clinical conditions. The main characteristics of the datasets are summarized in Table~\ref{tab:dataset}. 

\begin{table}[!ht]
\caption{properties of the public cone-beam CT (CBCT) datasets.\label{tab:dataset}}
\centering
\begin{tabular}{lccc}
\midrule
\toprule[1pt]
Dataset              &\#Patients  &\#Slices &Slice size  \\ \hline
\textit{LUNG TCIA} \cite{r13} &20 &6,721 & 512 × 512      \\
PELVIC TCIA \cite{clark2013cancer}  &58 &3,850 & 512 × 512        \\
\midrule
\bottomrule[1pt]
\end{tabular}
 \begin{tablenotes}
    \small
    \item[*]\textbf{Remark}: \#Patients and \#Slices denote the number of Patients, and the slices for each Patients, respectively.  
      \end{tablenotes}
\end{table}

For pre-processing, we treated the data as unpaired 2D CT/CBCT slices. All images were padded to a unified spatial resolution and normalized to the range \([-1,1]\) during training, and subsequently mapped to \([0,1]\) for evaluation. Peak signal-to-noise ratio (PSNR) \cite{poobathy2014edge} was computed after per-slice contrast normalization.

For quantitative evaluation, paired CBCT and pCT images from the lung cancer cohort were used, with pCT serving as the reference standard. Axial slices from CBCT, pCT, and synthesized images were compared under the same window setting to visually assess whether the proposed method improved image quality and anatomical fidelity. Quantitative performance was evaluated using mean absolute error (MAE) \cite{chai2014root}, root mean squared error (RMSE) \cite{chai2014root}, PSNR, and structural similarity index measure (SSIM) \cite{chen2011fast}.
 All competing methods were trained using unpaired CBCT and pCT images under the same unsupervised learning setting. 



\subsection{Performance comparison} \label{ssec:result}

We compared RefineCBCT with representative GAN-based and diffusion-based methods on the \textit{LUNG TCIA} and \textit{PELVIC TCIA} datasets, as summarized in Table~\ref{tab:main_results}. Overall, RefineCBCT achieved the most competitive performance, with particularly notable advantages in the error-based metrics MAE and RMSE, which directly reflect reconstruction accuracy.

\begin{table*}[!hbt]
    \centering
    \small
    \caption{Quantitative comparison of the image quality produced using various models.}
    \label{tab:main_results}
    \resizebox{\linewidth}{!}{
    \begin{threeparttable}
    \begin{tabular}{clccccc}
    \toprule[1pt]
    \midrule
Dataset &Methods & MAE ↓ & RMSE ↓ & PSNR ↑ & SSIM ↑\\ \hline
\multirow{9}{*}{\rotatebox{90}{\textit{LUNG TCIA}}}    
    &CycleGAN$^{+}$\cite{zhu2017unpaired}&36.503 & 97.512  & 26.499 & 0.848 \\
    &CUT$^{+}$ \cite{r5}&37.684 & 101.165 & 26.048 & 0.849 \\
    &DCLGAN$^{+}$ \cite{DCLGAN}& 32.503 & \underline{88.149}&27.234 & 0.876 \\
    &SynDiff$^{*}$ \cite{syndiff}& 34.791 & 94.493  & 26.713 & 0.858 \\
    &FGDM$^{*}$ \cite{FGDM}& 31.164& 88.817& 27.323 & 0.879 \\
    &DiffusionMBIR$^{*}$ \cite{chung2023solving}&\underline{26.874} &89.037 &\underline{28.819} &\underline{0.897} \\
    &WIA-LD2ND$^{+}$ \cite{zhao2024wia}&29.873 &90.871 &28.108 &0.880 \\ 
    &TLIR$^{*}$ \cite{li2025tlir}&29.403 &90.723 &28.096 &0.876 \\
    &\textbf{RefineCBCT}$^{*}$&\textbf{19.411} & \textbf{62.758} & \textbf{30.845} & \textbf{0.931} \\ 
    &Improvement (\%) & 7.463$\sim$17.092  & 25.391$\sim$38.407 & 2.026$\sim$4.797 &0.034$\sim$0.083 \\
\midrule
\bottomrule[1pt]
\multirow{9}{*}{\rotatebox{90}{\textit{PELVIC TCIA}}}    
    &CycleGAN$^{+}$\cite{zhu2017unpaired}&21.065&63.932&30.927&0.806 \\
    &CUT$^{+}$ \cite{r5}&28.167&82.429&32.143&0.763 \\
    &DCLGAN$^{+}$ \cite{DCLGAN}&22.072&68.315&32.708&0.833 \\
    &SynDiff$^{*}$ \cite{syndiff}&\underline{17.253} &\textbf{54.729} &34.476&0.828 \\
    &FGDM$^{*}$ \cite{FGDM}&19.011&67.113 &34.606 &\underline{0.842} \\
    &DiffusionMBIR$^{*}$ \cite{chung2023solving}&18.097 &55.861 &\underline{35.298} &0.830 \\
    &WIA-LD2ND$^{+}$ \cite{zhao2024wia}&19.683 &60.730 &35.103 &0.819 \\
    &TLIR$^{*}$ \cite{li2025tlir}&19.726 &66.351 &34.096 &0.822 \\
    &\textbf{RefineCBCT}$^{*}$&\textbf{14.905} &\underline{55.263} &\textbf{36.671} &\textbf{0.876} \\
    &Improvement (\%) &2.348$\sim$13.262 & -0.534$\sim$27.166 &1.373$\sim$5.744 &0.034$\sim$0.113 \\
\midrule
\bottomrule[1pt]
\end{tabular}
\begin{tablenotes}
    \small
    \item[*]\textbf{Remark}: $+$ and $*$ indicate that the corresponding method uses ResNet-9 and U-Net as the backbone, respectively. $\uparrow$($\downarrow$) indicates that the larger (smaller) the value, the better the performance; \textbf{Bold} and \underline{underlined} values indicate the best and second-best performance, respectively.
\end{tablenotes}
    \end{threeparttable}
    }
\end{table*}

\begin{figure*}[!ht]
    \centering
    \begin{subfigure}{\textwidth}
        \centering
        \includegraphics[width=\textwidth]{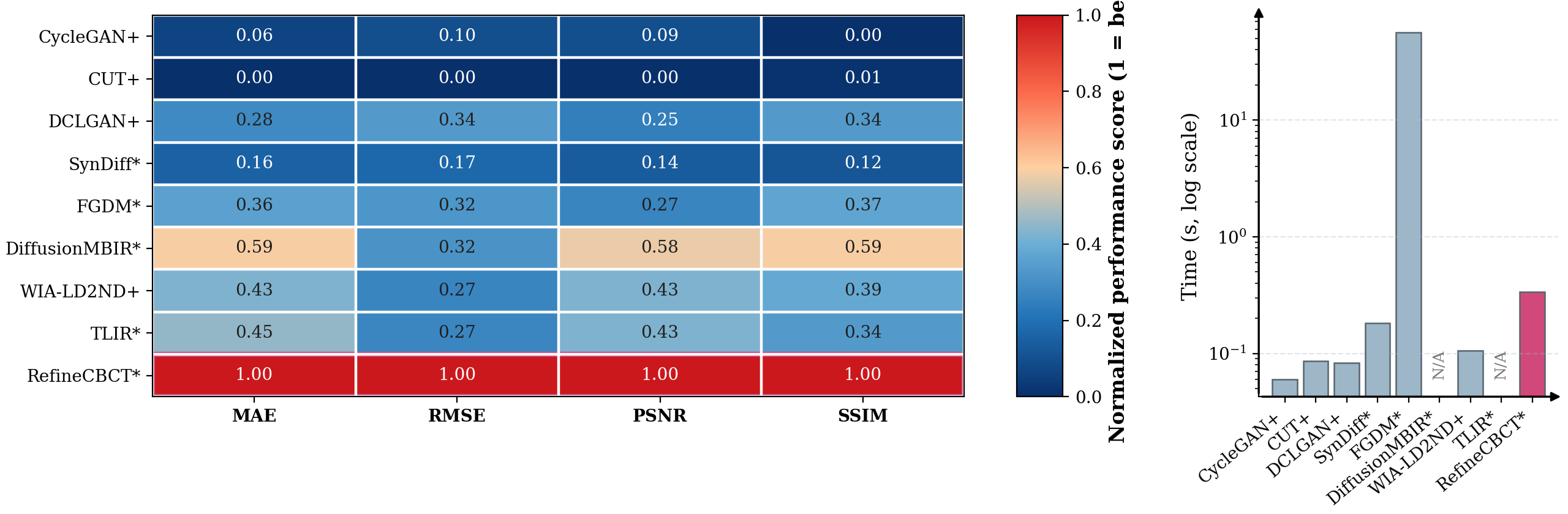}
        \caption{\small Result on LUNG TCIA dataset.}
    \end{subfigure}
    \vspace{-0.8em}
    \begin{subfigure}{\textwidth}
        \centering
        \includegraphics[width=\textwidth]{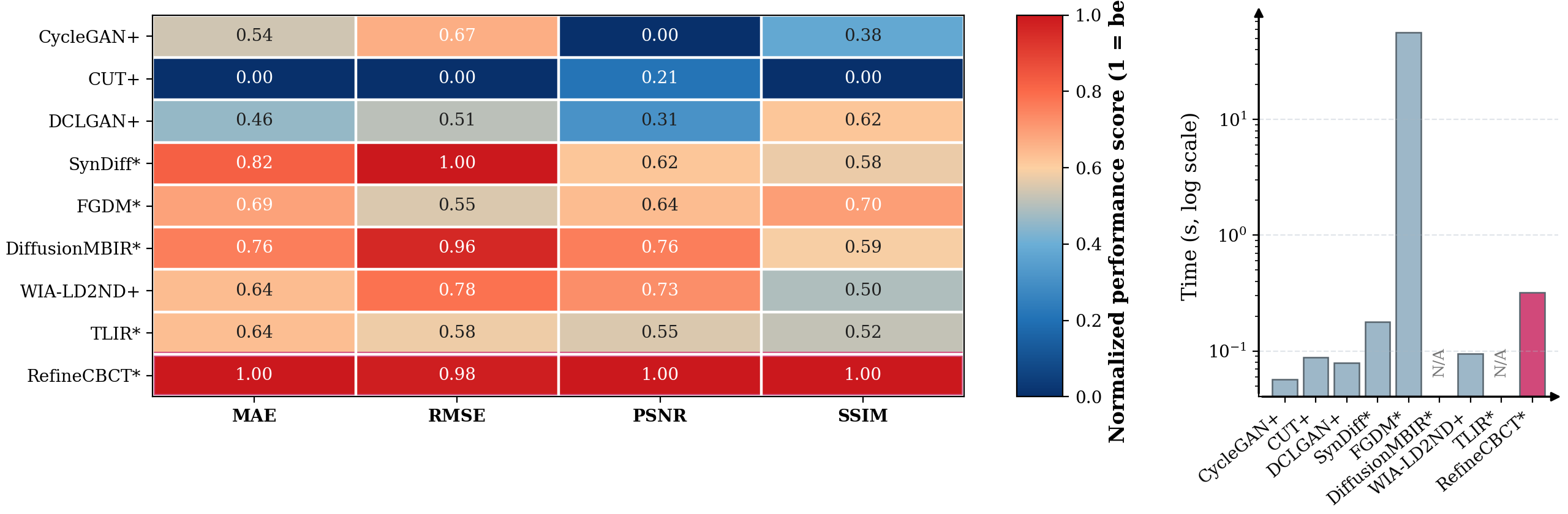}
        \caption{\small Result on the PELVIC TCIA dataset.}
    \end{subfigure}
    \vspace{-0.4em}
    \caption{Experimental analysis across various methods on LUNG TCIA and PELVIC TCIA datasets. For each dataset, the left panel shows normalized quality scores for MAE, RMSE, PSNR, and SSIM, and the right panel presents the corresponding inference time on a logarithmic scale.}
    \label{fig:main_results}
\end{figure*}

To provide a more intuitive comparison of overall performance, Fig.~\ref{fig:main_results} presents the normalized quality scores and inference times on the \textit{LUNG TCIA} and \textit{PELVIC TCIA} datasets. The heatmaps summarize the relative performance of each method in terms of MAE, RMSE, PSNR, and SSIM, while the bar plots show the corresponding inference time on a logarithmic scale. Combined with Table~\ref{tab:main_results}, these results show that RefineCBCT achieves the best overall trade-off between reconstruction quality and computational efficiency.

Combining Table~\ref{tab:main_results} and Fig.~\ref{fig:main_results}, we observe that RefineCBCT consistently outperforms existing GAN-based and diffusion-based methods in terms of reconstruction accuracy and image fidelity. On the \textit{LUNG TCIA} dataset, RefineCBCT achieved the best performance across all four metrics. In particular, the substantial reductions in MAE and RMSE highlight its clear advantage in suppressing severe artifacts and reducing pixel-wise intensity deviations in challenging lung cases. Meanwhile, the highest PSNR (30.845 dB) and SSIM (0.931) further indicate improved image fidelity and structural preservation.

On the \textit{PELVIC TCIA} dataset, RefineCBCT again achieved the best MAE, PSNR, and SSIM, while obtaining the second-best RMSE with only a marginal gap from the best result. Although SynDiff produced a slightly lower RMSE (54.729 vs.\ 55.263), RefineCBCT delivered more balanced and consistently stronger performance across the remaining metrics. These results demonstrate that RefineCBCT not only reduces reconstruction errors effectively, but also preserves image content and anatomical structure more reliably across different anatomical sites..

\subsection{Qualitative evaluation}

\begin{figure*}[!htb]
\begin{subfigure}{1\textwidth}
    \centering
    \includegraphics[width=\textwidth]{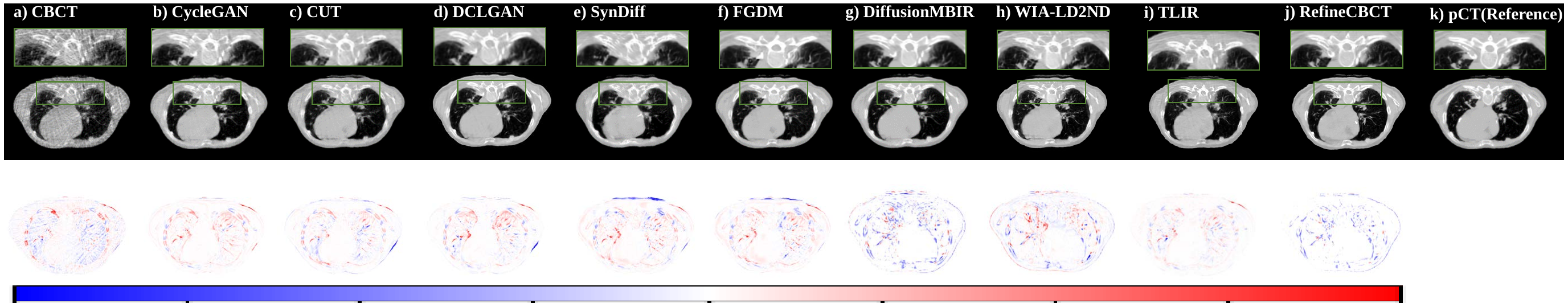}
    \caption{\small{Result on LUNG TCIA dataset.}}
\end{subfigure}%
\hspace{-0.5cm}
\begin{subfigure}{1\textwidth}
    \centering
    \includegraphics[width=\textwidth]{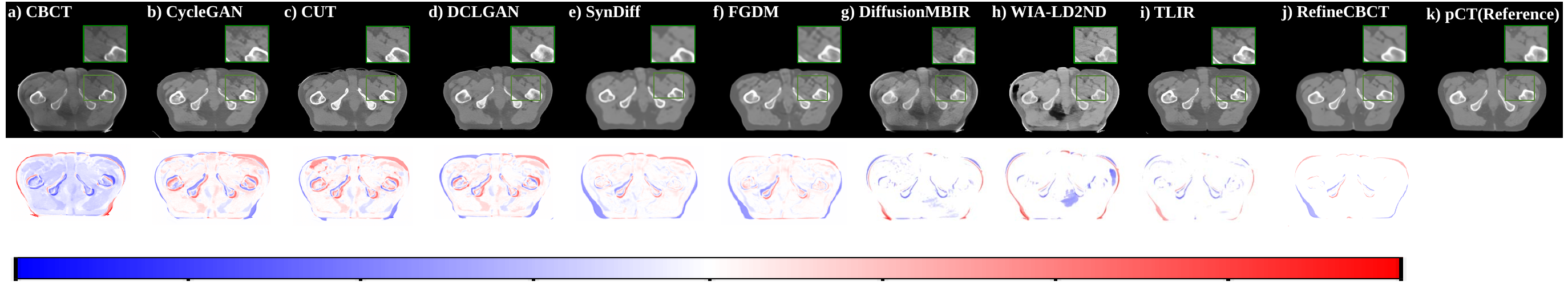}
    \caption{\small{Result on PELVIC TCIA dataset.}}
\end{subfigure}%
  \caption{Qualitative comparison of RefineCBCT against compared methods. For each subfigure, the middle row shows representative axial slices for all methods, the top row provides magnified views of the green boxed regions for fine-structure inspection, and the bottom row shows difference maps with respect to pCT. The pCT is used as the reference for visual comparison. Display window: [-1000, 1000] HU.}
   \label{fig: visualization result}
\end{figure*}

Figure~\ref{fig: visualization result} further compares RefineCBCT with SOTA methods using zoomed-in regions and difference maps relative to pCT. on \textit{LUNG TCIA} dataset (Figure~\ref{fig: visualization result}a), where artifacts are more severe, RefineCBCT suppresses both global shading and local streak artifacts more effectively than the competing methods. The enlarged regions show clearer tissue boundaries and better recovery of fine structures, while the difference maps show smaller residual errors. On the \textit{PELVIC TCIA} dataset (Figure~\ref{fig: visualization result}b), RefineCBCT also produces cleaner images with more uniform soft tissue and fewer visible artifacts. Although the improvement is less pronounced than in \textit{LUNG TCIA}, the results remain visually closer to pCT than those of the comparison methods.

\begin{figure*}[!ht]
\centering
\begin{subfigure}[t]{0.24\textwidth}
    \centering
    \includegraphics[width=\linewidth]{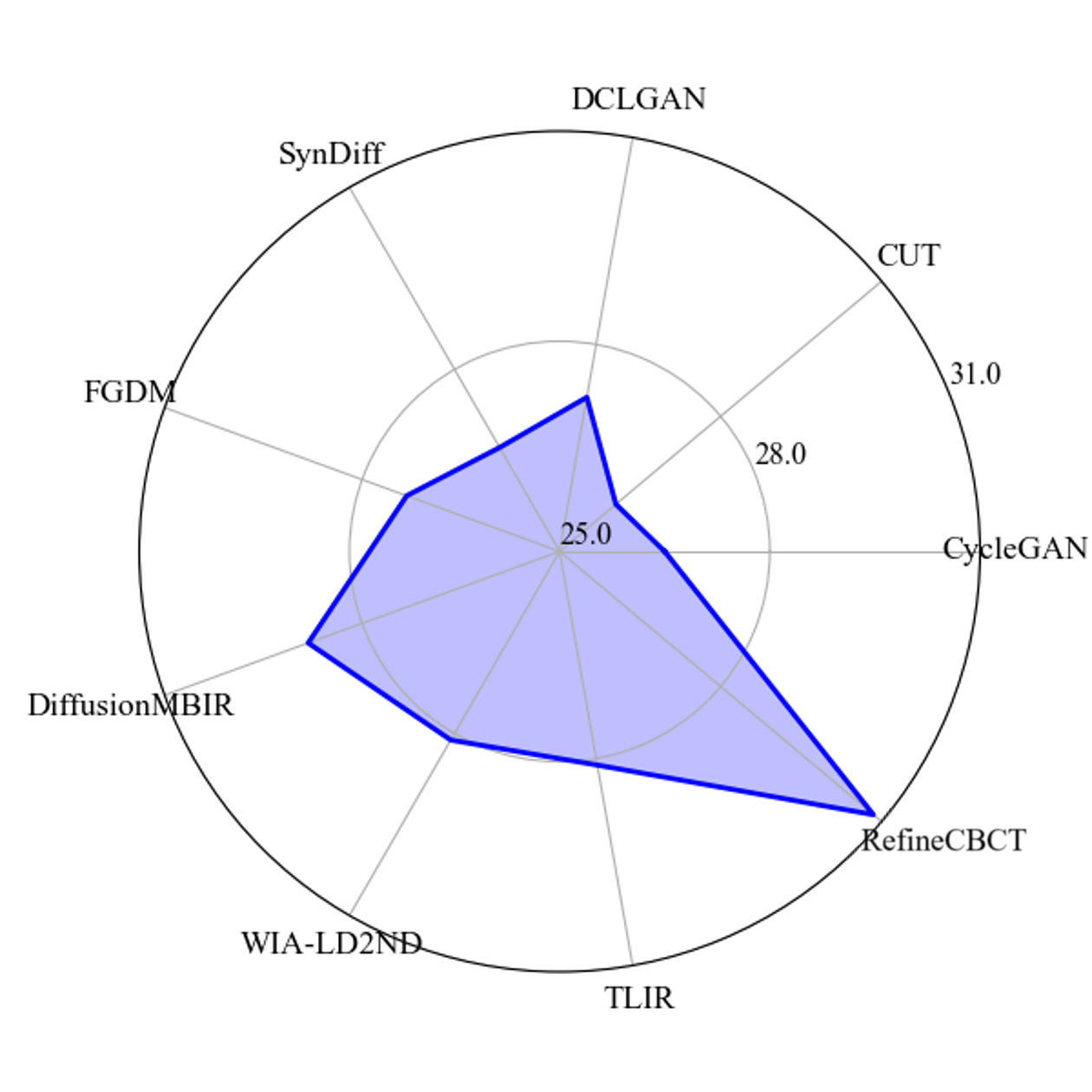}
    \caption{\small LUNG PSNR}
\end{subfigure}\hfill
\begin{subfigure}[t]{0.24\textwidth}
    \centering
    \includegraphics[width=\linewidth]{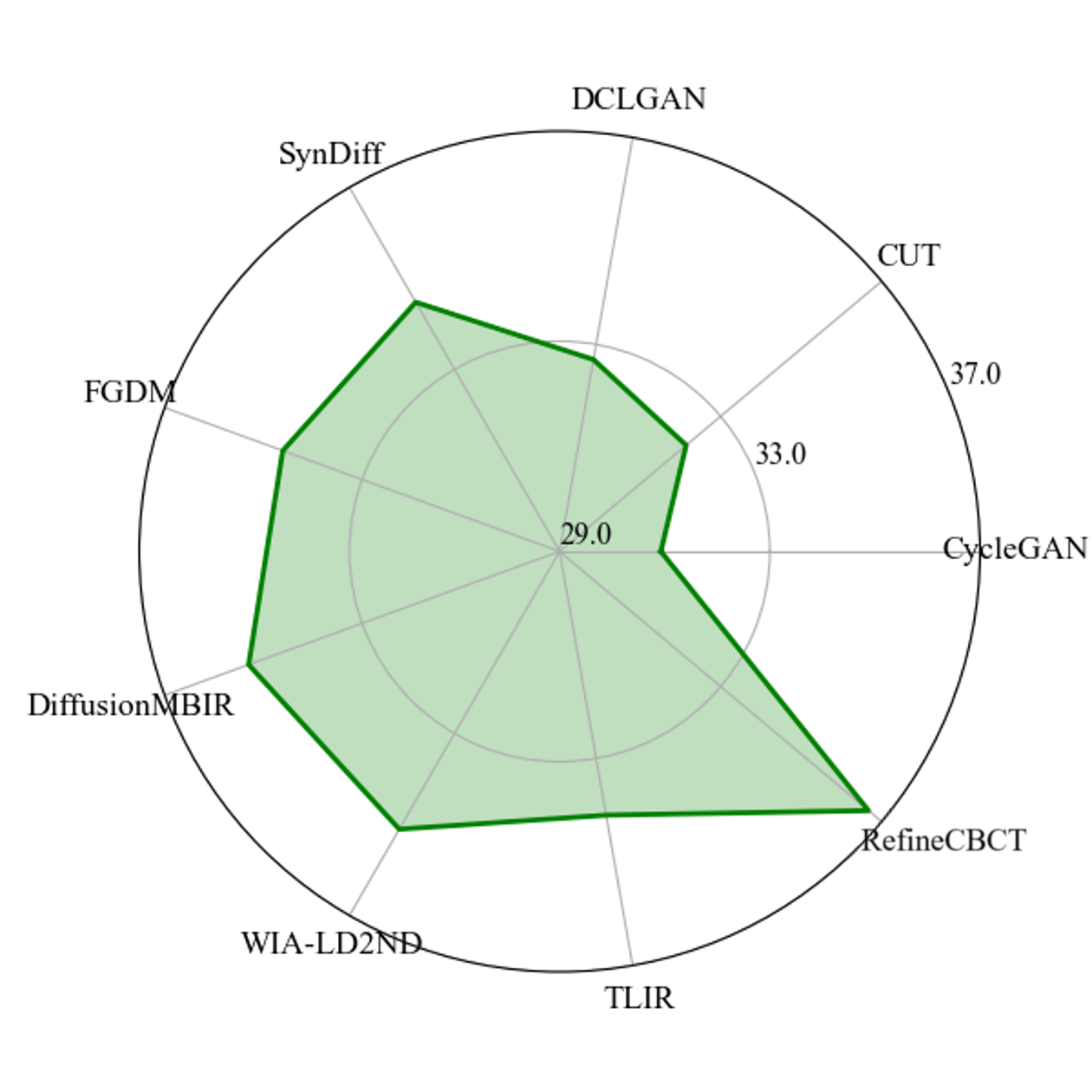}
    \caption{\small PELVIC PSNR}
\end{subfigure}\hfill
\begin{subfigure}[t]{0.24\textwidth}
    \centering
    \includegraphics[width=\linewidth]{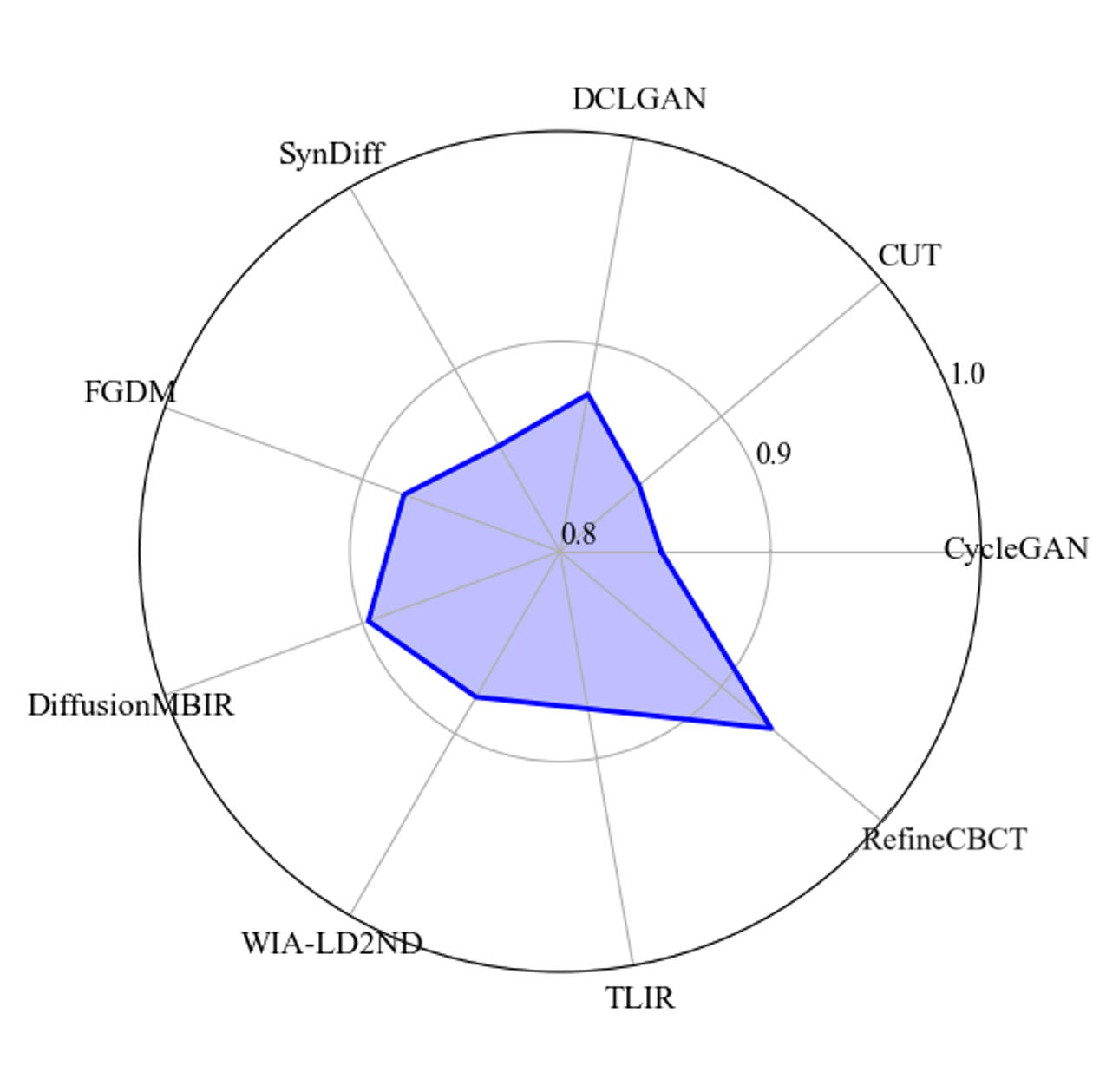}
    \caption{\small LUNG SSIM}
\end{subfigure}\hfill
\begin{subfigure}[t]{0.24\textwidth}
    \centering
    \includegraphics[width=\linewidth]{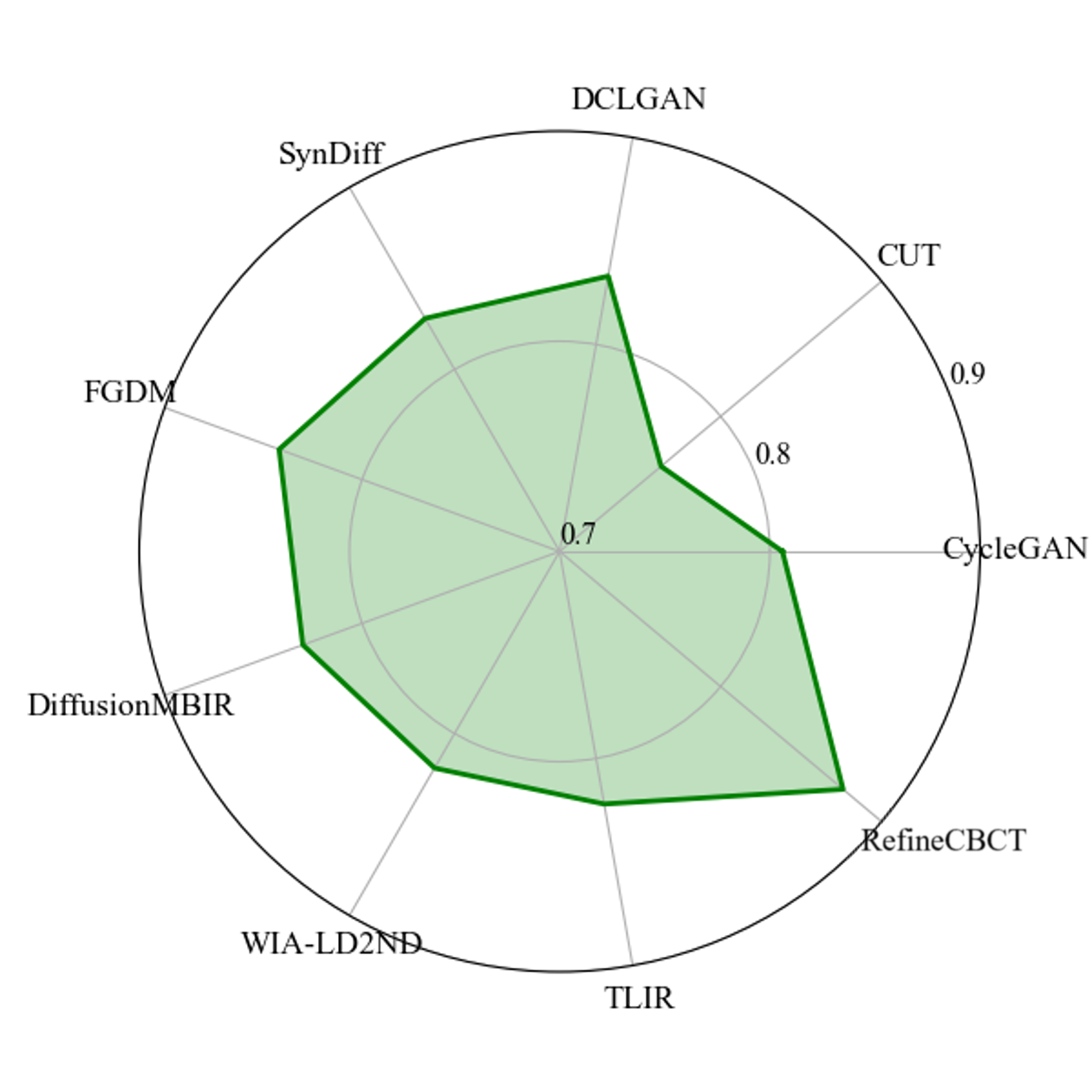}
    \caption{\small PELVIC SSIM}
\end{subfigure}

\caption{Radar-plot comparison of RefineCBCT and baseline methods on two public datasets. Larger values indicate better performance.}
\label{fig:radar_analysis}
\end{figure*}

Figure~\ref{fig:radar_analysis} summarizes the quantitative comparison between RefineCBCT and the baseline methods in terms of PSNR and SSIM. Across all four radar plots, both the visual results and the quantitative analysis show that RefineCBCT achieves a favorable balance among artifact reduction, structural preservation, and CT-value fidelity across different anatomical sites, demonstrating superior or near-superior performance on both datasets. The advantage is more pronounced on \textit{LUNG TCIA}, which is consistent with the stronger visual improvements observed in cases with heavier artifacts. On \textit{PELVIC TCIA}, the gains are smaller but remain consistent.


\subsection{Quantitative CT value assessment}

\begin{figure*}[!htb]
\centering
\begin{subfigure}{0.48\textwidth}
    \centering
    \includegraphics[width=\textwidth]{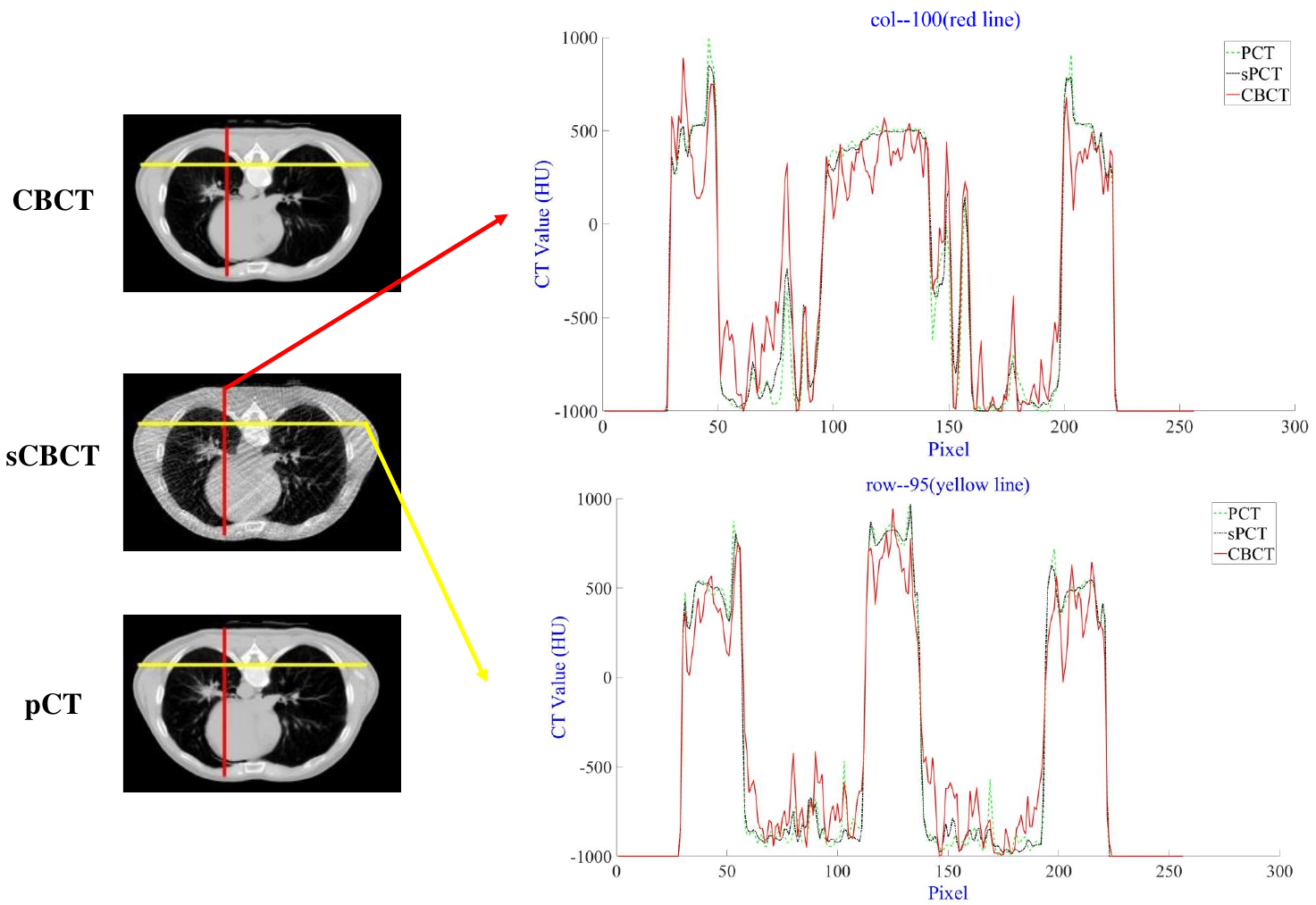}
    \caption{\small{CT value on LUNG TCIA}}
\end{subfigure}%
\begin{subfigure}{0.48\textwidth}
    \centering
    \includegraphics[width=\textwidth]{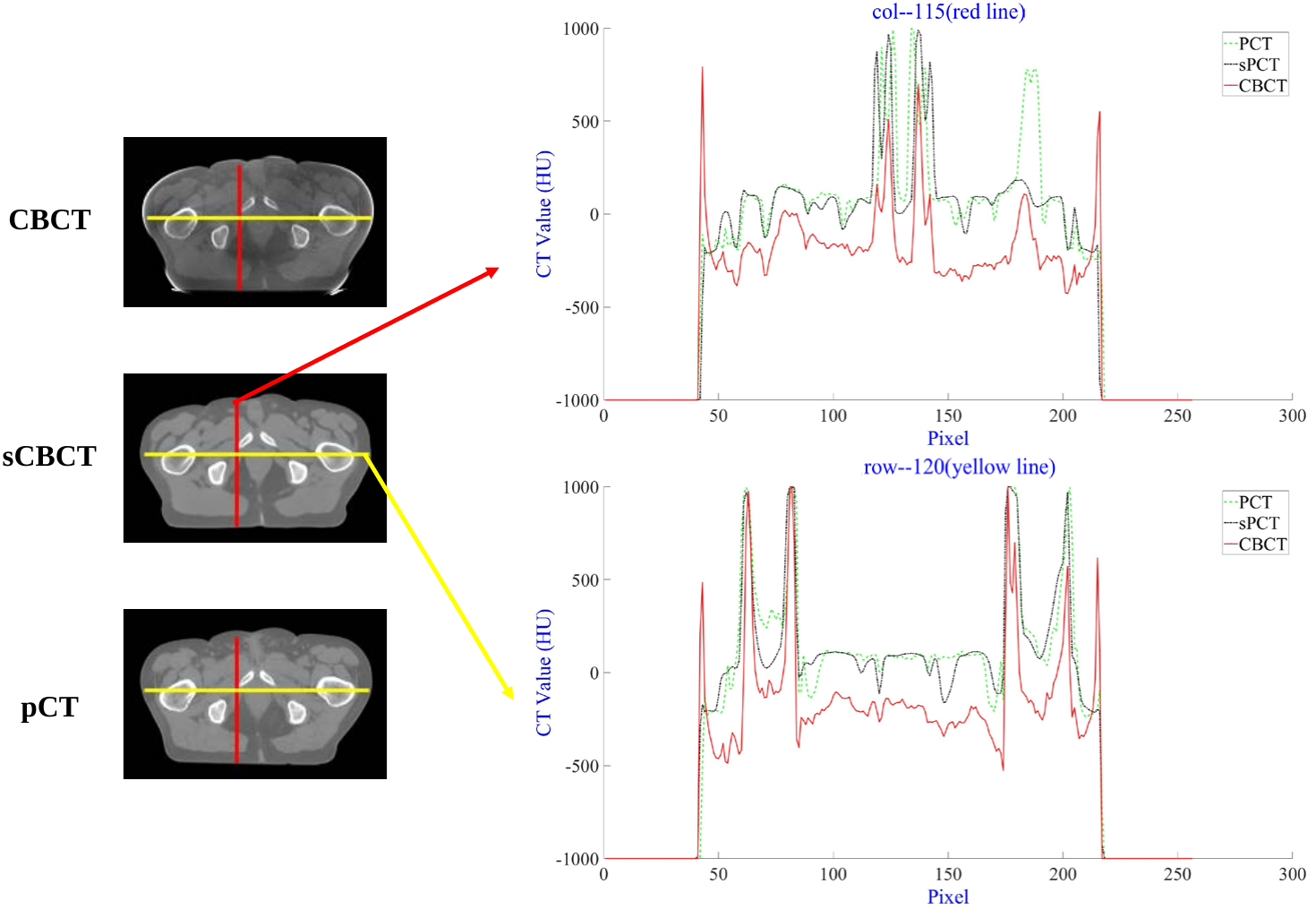}
    \caption{\small{CT value on PELVIC TCIA}}
\end{subfigure}
\caption{Line-profile analysis of HU consistency for RefineCBCT on two public datasets. For each subfigure, the left panel shows matched axial slices of CBCT, sCBCT, and pCT, and the right panel plots HU profiles sampled along the red and yellow lines. (a) On LUNG TCIA, the red line (Line~100) mainly traverses soft tissue, whereas the yellow line (Line~95) crosses both bone and soft tissue. (b) On PELVIC TCIA, the red line (Line~115) mainly traverses soft tissue, whereas the yellow line (Line~120) crosses both bone and soft tissue. In both datasets, sCBCT more closely follows pCT and shows reduced fluctuations relative to CBCT. Display window: [-1000, 1000] HU.}
\label{fig:HU value distributions}
\end{figure*}

To evaluate CT-number fidelity after correction, we analyzed HU consistency using line profiles and region-based distributions. Figure~\ref{fig:HU value distributions} shows representative HU line profiles on \textit{LUNG TCIA} and \textit{PELVIC TCIA} datasets, sampled along two predefined lines on matched axial slices of CBCT, sCBCT, and pCT.

\begin{figure*}[!ht]
\centering
\begin{subfigure}{0.48\textwidth}
    \centering
    \includegraphics[width=\textwidth]{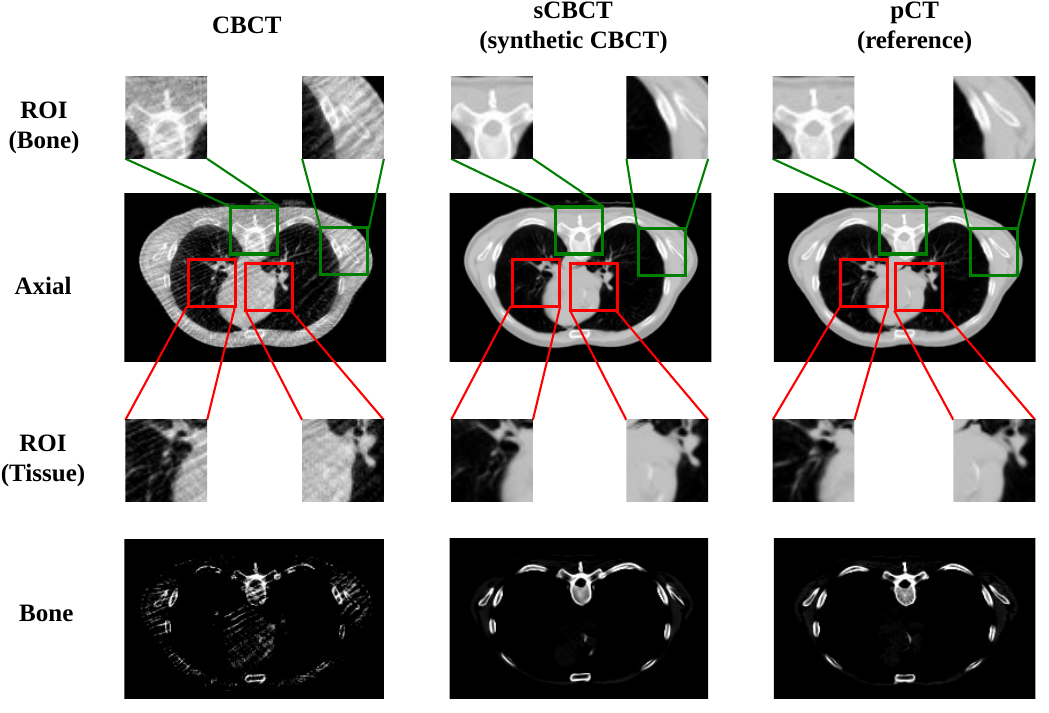}
    \caption{\small{Distribution of ROIs on LUNG TCIA.}}
\end{subfigure}%
\hfill
\begin{subfigure}{0.48\textwidth}
    \centering
    \includegraphics[width=\textwidth]{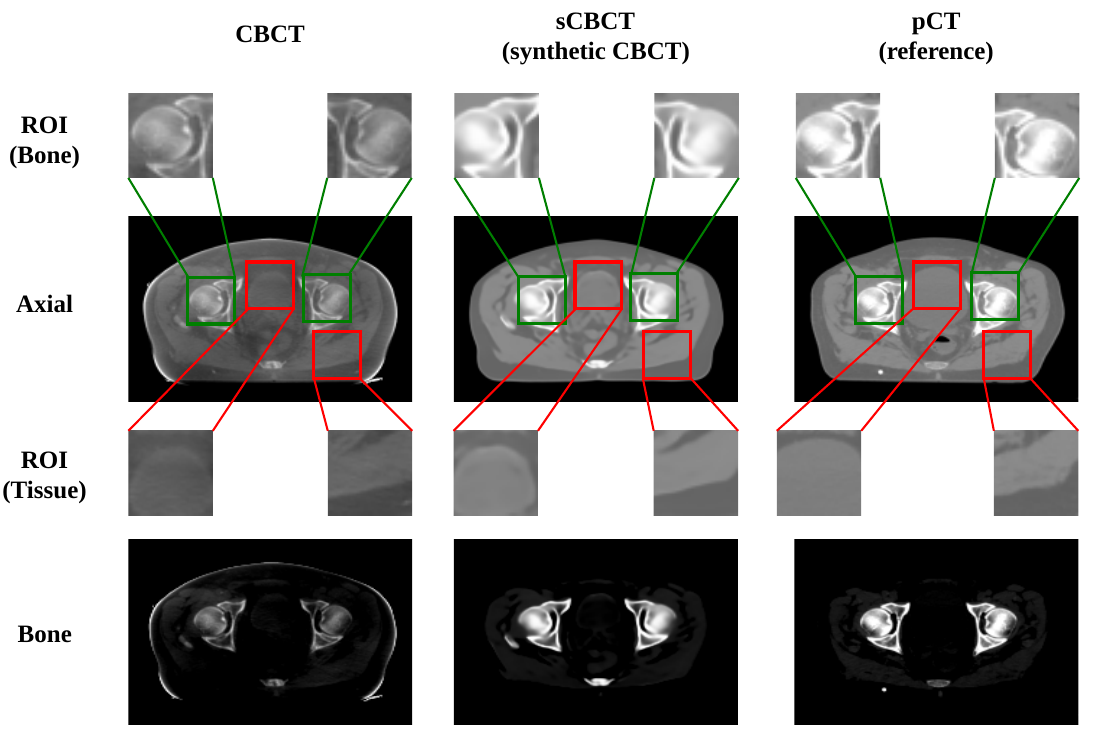}
    \caption{\small{Distribution of ROIs on PELVIC TCIA.}}
\end{subfigure}
\begin{subfigure}{0.48\textwidth}
    \centering
    \includegraphics[width=\textwidth]{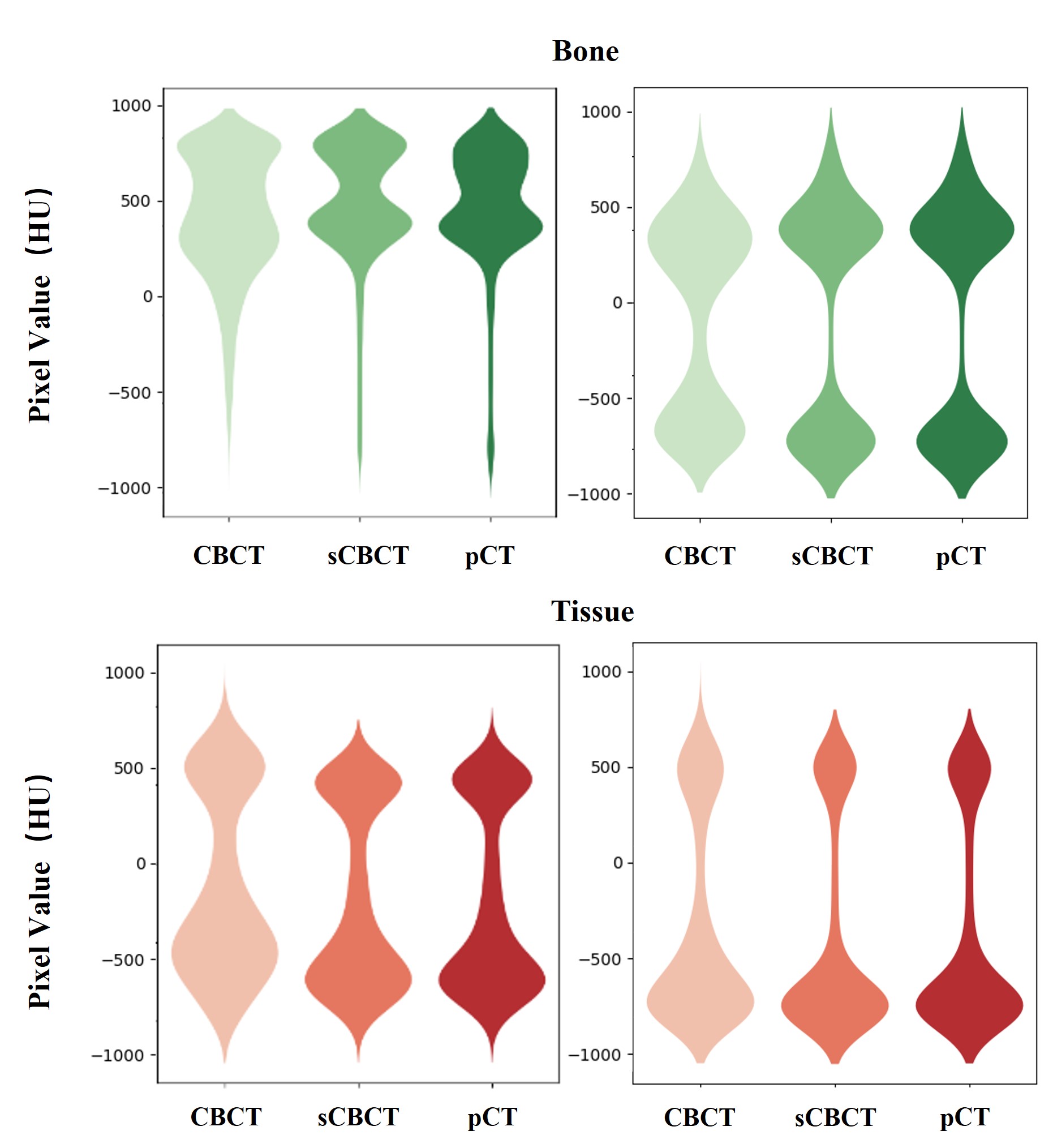}
    \caption{\small{Distribution of CT value on LUNG TCIA.}}
\end{subfigure}%
\hfill
\begin{subfigure}{0.48\textwidth}
    \centering
    \includegraphics[width=\textwidth]{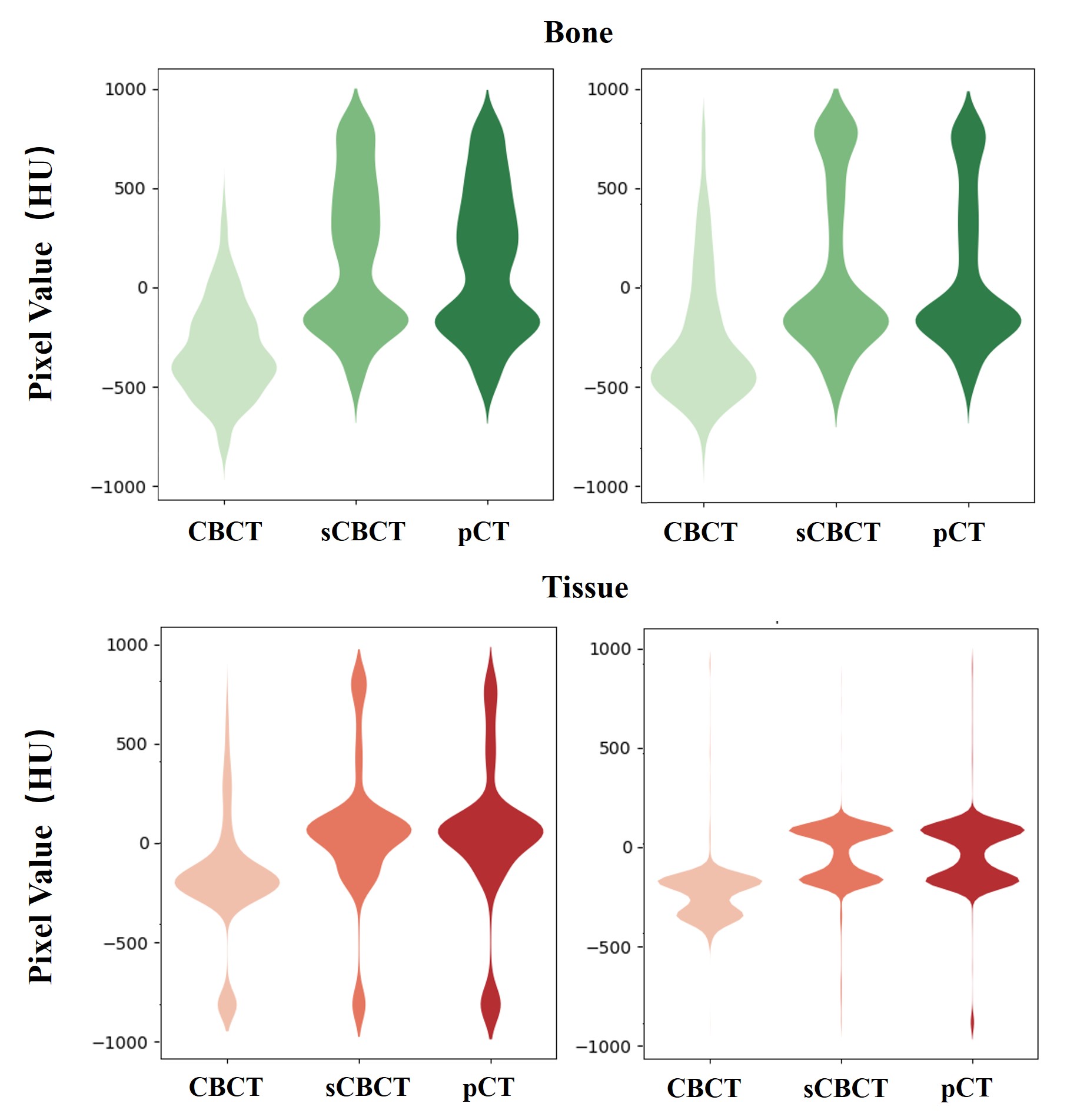}
    \caption{\small{Distribution of CT value on PELVIC TCIA.}}
\end{subfigure}

\caption{ROI placement and regional HU distribution analysis on two public datasets. (a) and (b) show representative bone and soft-tissue ROIs placed at corresponding anatomical locations in CBCT, sCBCT, and pCT. Green boxes denote bone ROIs and red boxes denote soft-tissue ROIs. From top to bottom, the rows show ROI (Bone), axial slice, ROI (Tissue), and a high-contrast bone window. The display windows for ROI (Bone), axial slice, and ROI (Tissue) are [-1000, 1000] HU, and that for the bottom bone row is [400, 750] HU. (c) and (d) show violin plots of HU distributions within the selected bone and soft-tissue ROIs for LUNG TCIA and \textit{PELVIC TCIA}, respectively. Compared with CBCT, sCBCT exhibits distributions that are narrower and closer to those of pCT, indicating reduced bias and variance in both bone and soft-tissue regions.}
\label{fig:HU value distributions of ROI}
\end{figure*}

On \textit{LUNG TCIA} dataset (Figure~\ref{fig:HU value distributions}(a)), the sCBCT profiles follow the pCT curves more closely than the original CBCT. This is observed for both the soft-tissue line and the line crossing bone and soft tissue, where sCBCT shows fewer fluctuations and clearer intensity transitions at tissue boundaries. On the \textit{PELVIC TCIA} dataset (Figure~\ref{fig:HU value distributions}(b)), similar trends are observed. The sCBCT profiles are smoother than CBCT and better match the pCT curves, especially in regions containing both bone and soft tissue. Overall, the line-profile analysis shows that RefineCBCT improves local HU accuracy and reduces intensity fluctuations on both datasets.

Figure~\ref{fig:HU value distributions of ROI} further evaluates regional HU fidelity using bone and soft-tissue ROIs placed at corresponding anatomical locations in CBCT, sCBCT, and pCT. Panels (a) and (b) show the ROI placement for the \textit{LUNG TCIA} and \textit{PELVIC TCIA} datasets, and panels (c) and (d) show the corresponding violin plots of HU distributions.
For both datasets, the HU distributions of \textbf{sCBCT} are closer to those of \textbf{pCT} than those of the original \textbf{CBCT}. In both bone and soft-tissue ROIs, CBCT shows broader and more biased distributions, while sCBCT is narrower and better aligned with pCT. This indicates that RefineCBCT improves HU consistency in both high-density and soft-tissue regions.

\subsection{Ablation study}

We conducted ablation experiments on the \textit{LUNG TCIA} and \textit{PELVIC TCIA} datasets to evaluate the contribution of each component in RefineCBCT, as reported in Tables~\ref{tab:ablation_lung} and \ref{tab:ablation_pelvic}.

\begin{table}[htbp]
\centering
\small
\caption{Ablation study of each component on LUNG TCIA dataset.}
\label{tab:ablation_lung}
\begin{tabular}{lccc}
\midrule
\toprule[1pt]
Method variant & Inference time (s) $\downarrow$ & PSNR (dB) $\uparrow$ & SSIM $\uparrow$ \\
\midrule
Diffusion backbone                    & \textbf{0.301} & 21.152 & 0.451 \\
Diffusion backbone + Generation       & 0.312          & 26.394 & 0.801 \\
Full model                            & 0.342          & \textbf{30.473} & \textbf{0.915} \\
\midrule
\bottomrule[1pt]
\end{tabular}
\vspace{2pt}
\small
\item[*]\textbf{Remark:}
The full model includes the diffusion backbone, pseudo-label generation module, and cycle-consistency loss (C.C.L). $\uparrow$ and $\downarrow$ indicate that higher and lower values are better, respectively.
\end{table}

Using only the \emph{Diffusive} module gives the lowest inference time, but also the weakest image quality. After adding the pseudo-label \emph{Generation} module, performance improves substantially, with PSNR increasing to 26.394 dB and SSIM to 0.801 on \textit{LUNG TCIA}, and to 31.259 dB and 0.775 on \textit{PELVIC TCIA}, while the increase in inference time remains small. This result highlights the importance of the generation module, which enables effective unpaired training by producing pseudo-CBCT images from CT data. In this way, the model can learn artifact correction without relying on strictly paired CBCT--CT datasets, while maintaining structural consistency during training.

\begin{table}[htbp]
\centering
\small
\caption{Ablation study of each component on PELVIC TCIA dataset.}
\label{tab:ablation_pelvic}
\begin{tabular}{cccccc}
\midrule
\toprule[1pt]
Diffusive & Generation & C.C.L & Inference time (s) $\downarrow$ & PSNR (dB) $\uparrow$ & SSIM $\uparrow$ \\
\midrule
$\checkmark$ &  &  & \textbf{0.297} & 25.648 & 0.539 \\
$\checkmark$ & $\checkmark$ &  & 0.306 & 31.259 & 0.775 \\
$\checkmark$ & $\checkmark$ & $\checkmark$ & 0.322 & \textbf{35.941} & \textbf{0.848} \\
\midrule
\bottomrule[1pt]
\end{tabular}
\vspace{2pt}
\small
\item[*]\textbf{Remark:}
Diffusive, Generation, and C.C.L denote the pseudo-label guided diffusion module, pseudo-label generation module, and cycle-consistency loss, respectively. $\uparrow$ ($\downarrow$) indicates that higher (lower) values are better.
\end{table}

Adding the cycle-consistency loss (\emph{C.C.L}) further improves performance and yields the best results on both datasets. The full model reaches 30.473 dB PSNR and 0.915 SSIM on \textit{LUNG TCIA}, and 35.941 dB PSNR and 0.848 SSIM on \textit{PELVIC TCIA}, with only a modest additional increase in inference time. These results indicate that cycle-consistency loss plays an important role in stabilizing the mapping between CT and pseudo-CBCT domains. By encouraging the transformed image to preserve recoverable anatomical information, it helps reduce structural distortion and improves the reliability of unpaired learning.

\subsection{Impact on downstream tasks}

\begin{table*}[!hbt]
    \centering
    \small
    \caption{Quantitative comparison of downstream object detection performance produced using various models.}
    \label{tab:od_results}
    \begin{threeparttable}
    \begin{tabular}{clcccc}
    \toprule[1pt]
    \midrule
Dataset & Methods & mAP $\uparrow$ & $\Delta$ mAP & IoU $\uparrow$ & $\Delta$ IoU \\ \hline
\multirow{10}{*}{\rotatebox{90}{\textit{LUNG TCIA}}}
    & Original CBCT & 0.529 & -- & 0.189 & -- \\
    & CycleGAN$^{+}$\cite{zhu2017unpaired} & 0.617 & 0.088 & 0.193 & 0.004 \\
    & CUT$^{+}$ \cite{r5} & 0.607 & 0.078 & 0.279 & 0.090 \\
    & DCLGAN$^{+}$ \cite{DCLGAN} & 0.648 & 0.119 & 0.518 & 0.329 \\
    & SynDiff$^{*}$ \cite{syndiff} & 0.670 & 0.141 & 0.601 & 0.412 \\
    & FGDM$^{*}$ \cite{FGDM} & 0.660 & 0.131 & 0.505 & 0.316 \\
    & DiffusionMBIR$^{*}$ \cite{chung2023solving} & 0.661 & 0.132 & 0.696 & 0.507 \\
    & WIA-LD2ND$^{+}$ \cite{zhao2024wia} & 0.652 & 0.123 & 0.578 & 0.389 \\
    & TLIR$^{*}$ \cite{li2025tlir} & \underline{0.750} & \underline{0.221} & \underline{0.699} & \underline{0.510} \\
    & \textbf{RefineCBCT}$^{*}$ & \textbf{0.779} & \textbf{0.250} & \textbf{0.735} & \textbf{0.546} \\
\midrule
\multirow{10}{*}{\rotatebox{90}{\textit{PELVIC TCIA}}}
    & Original CBCT & 0.799 & -- & 0.559 & -- \\
    & CycleGAN$^{+}$\cite{zhu2017unpaired} & 0.830 & 0.031 & 0.674 & 0.115 \\
    & CUT$^{+}$ \cite{r5} & 0.875 & 0.076 & 0.744 & 0.185 \\
    & DCLGAN$^{+}$ \cite{DCLGAN} & 0.895 & 0.096 & \underline{0.820} & \underline{0.261} \\
    & SynDiff$^{*}$ \cite{syndiff} & 0.912 & 0.113 & 0.789 & 0.230 \\
    & FGDM$^{*}$ \cite{FGDM} & 0.928 & 0.129 & 0.793 & 0.234 \\
    & DiffusionMBIR$^{*}$ \cite{chung2023solving} & \textbf{0.941} & \textbf{0.142} & 0.800 & 0.241 \\
    & WIA-LD2ND$^{+}$ \cite{zhao2024wia} & 0.905 & 0.106 & 0.779 & 0.220 \\
    & TLIR$^{*}$ \cite{li2025tlir} & 0.931 & 0.132 & 0.799 & 0.240 \\
    & \textbf{RefineCBCT}$^{*}$ & \underline{0.939} & \underline{0.140} & \textbf{0.830} & \textbf{0.271} \\
\midrule
\bottomrule[1pt]
\end{tabular}
\begin{tablenotes}
    \small
    \item[*]\textbf{Remark}: $+$ and $*$ indicate that the corresponding method uses ResNet-9 and U-Net as the backbone, respectively. $\uparrow$ indicates that larger values correspond to better performance. \textbf{Bold} and \underline{underlined} values indicate the best and second-best results, respectively.
\end{tablenotes}
    \end{threeparttable}
\end{table*}

We further evaluated whether artifact correction improves downstream analysis using YOLOv8 for object detection. In image-guided radiotherapy, CBCT is used not only for visual assessment but also for automated tasks such as target localization and organ-at-risk identification. Following prior work, the original CBCT images were reconstructed using FDK \cite{feldkamp1984practical}. We then performed object detection on the corrected images. In Table~\ref{tab:od_results}, \(\Delta\)mAP and \(\Delta\)IoU denote the performance gains relative to the original CBCT.

As shown in Table~\ref{tab:od_results}, artifact correction generally improved detection performance compared with the original CBCT images, although the degree of improvement varied across methods and datasets. On the LUNG TCIA dataset, RefineCBCT achieved the best results on all four metrics, increasing mAP from 0.529 to 0.779 and IoU from 0.189 to 0.735, with gains of 0.250 and 0.546, respectively. On the PELVIC TCIA dataset, RefineCBCT achieved the best IoU and \(\Delta\)IoU, and the second-best mAP and \(\Delta\)mAP, demonstrating consistently strong downstream detection performance.

\begin{figure*}[!htb]
\centering
\begin{subfigure}[t]{0.47\textwidth}
    \centering
    \includegraphics[height=0.77\textheight]{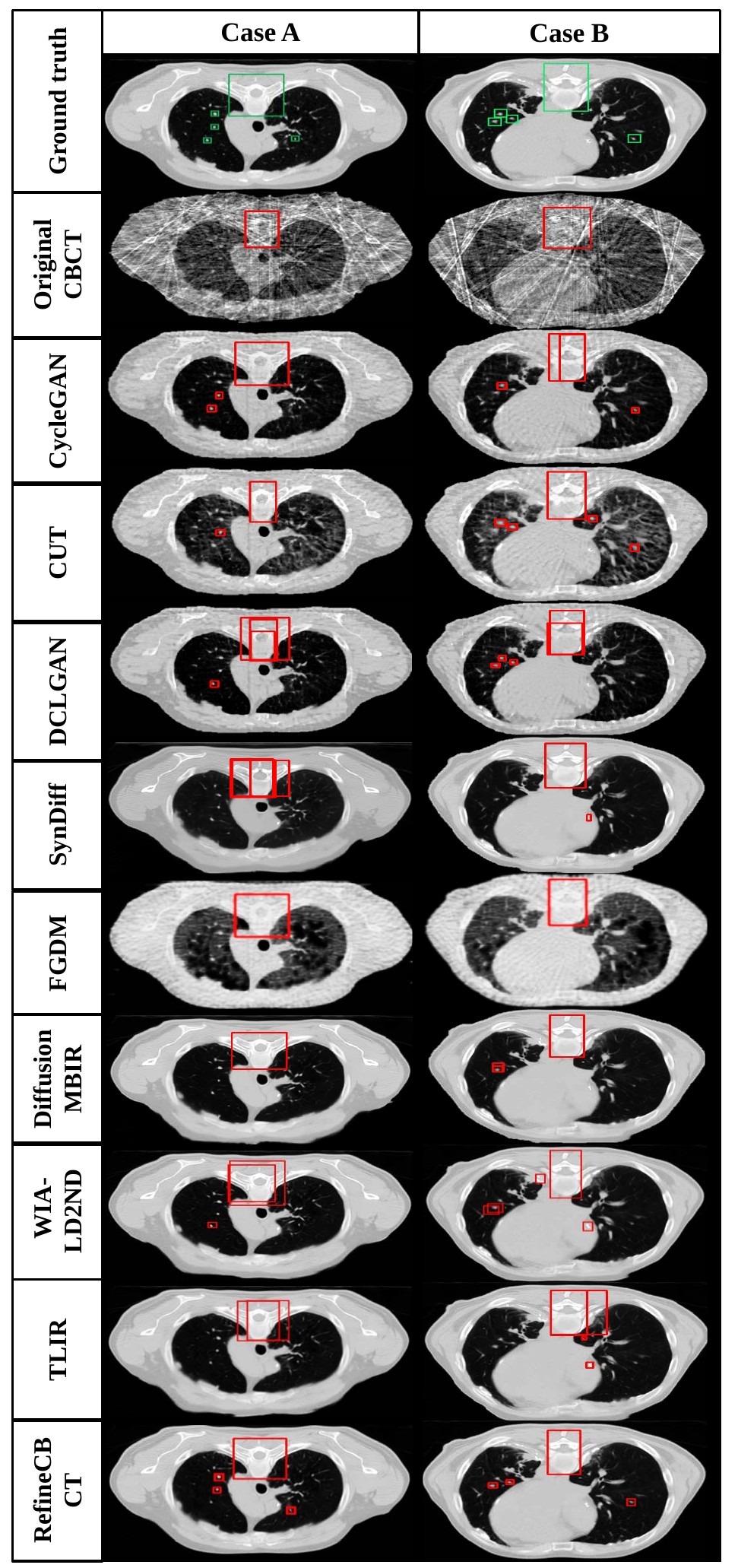}
    \caption{LUNG TCIA}
    \label{fig:od_lung}
\end{subfigure}
\hfill
\begin{subfigure}[t]{0.47\textwidth}
    \centering
    \includegraphics[height=0.77\textheight]{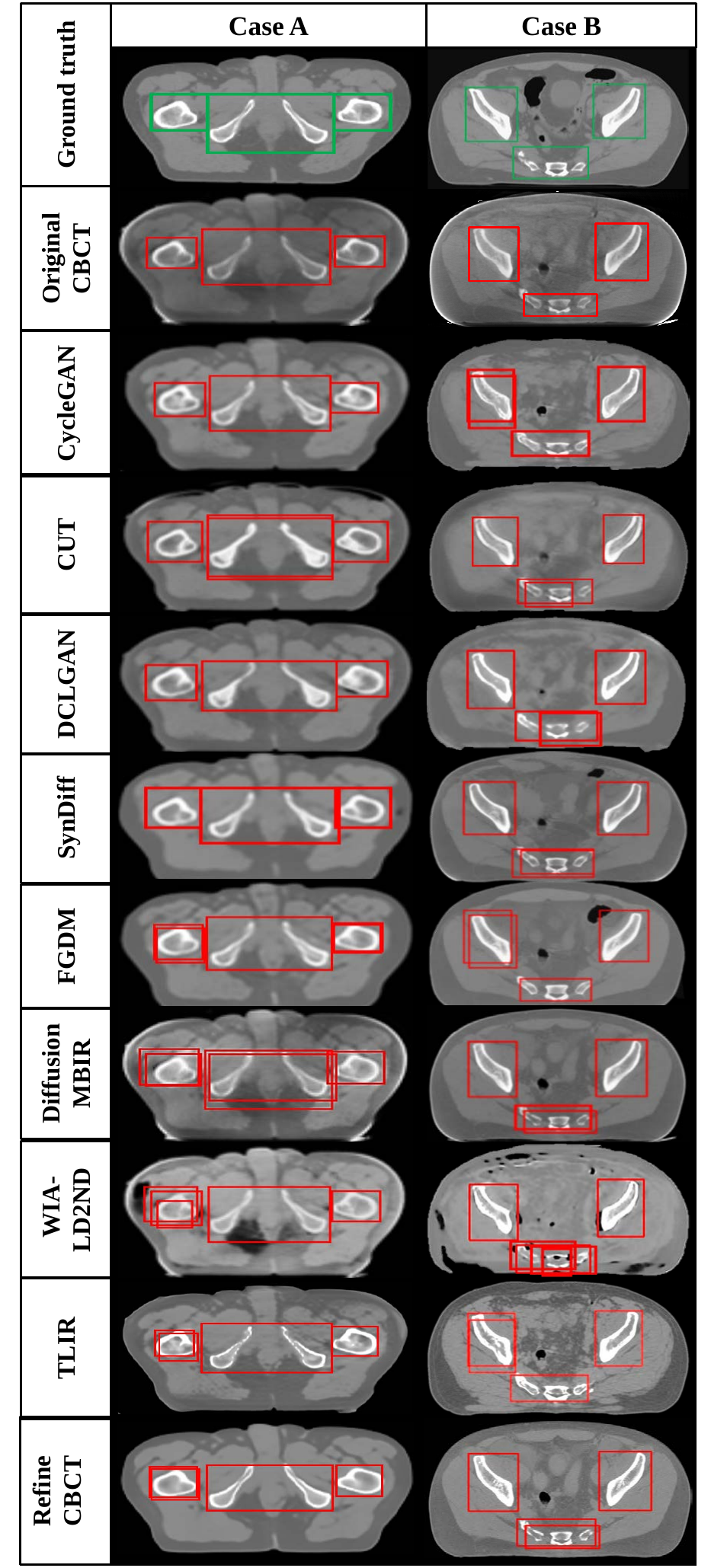}
    \caption{PELVIC TCIA}
    \label{fig:od_pelvic}
\end{subfigure}
\caption{Qualitative object detection results on the LUNG and PELVIC TCIA datasets. Red boxes indicate predictions and green boxes indicate the reference targets.}
\label{fig:od_results}
\end{figure*}

The qualitative results further support these findings. In the lung cases shown in Fig.~\ref{fig:od_lung}, the detection task involves both a relatively large target region in the spine and small rounded nodular structures in the lung. In the original CBCT images, undersampling-induced streak artifacts severely degrade image quality, leading to poor identification of both target types. Most comparison methods improve the overall depiction of the spine, but their recovery of the small nodular structures remains limited. In contrast, RefineCBCT better preserves these subtle small-scale structures while maintaining the morphology of the larger target, which is consistent with its superior quantitative detection performance. This suggests that the proposed method is more effective in recovering fine anatomical details that are critical for small-target recognition.

For the pelvic cases shown in Fig.~\ref{fig:od_pelvic}, the original CBCT images are substantially affected by scatter-induced artifacts, which increase noise, reduce overall intensity, and lower soft-tissue-to-bone contrast. As a result, the boundary contrast between target regions and surrounding tissues is weakened, making detection more difficult. After correction, the images show clearer structural contrast and reduced noise, with improved separation between bone and soft tissue. These changes are favorable for target recognition and are reflected in the improved detection metrics. Consistent with this observation, RefineCBCT achieved the highest IoU (0.830) and \(\Delta\)IoU (0.271), while remaining competitive in mAP.


\subsection{2D processing and volumetric consistency}
\label{ssec:volumetric_consistency}

\begin{figure*}[!htb]
    \centering
    \includegraphics[width=0.98\linewidth]{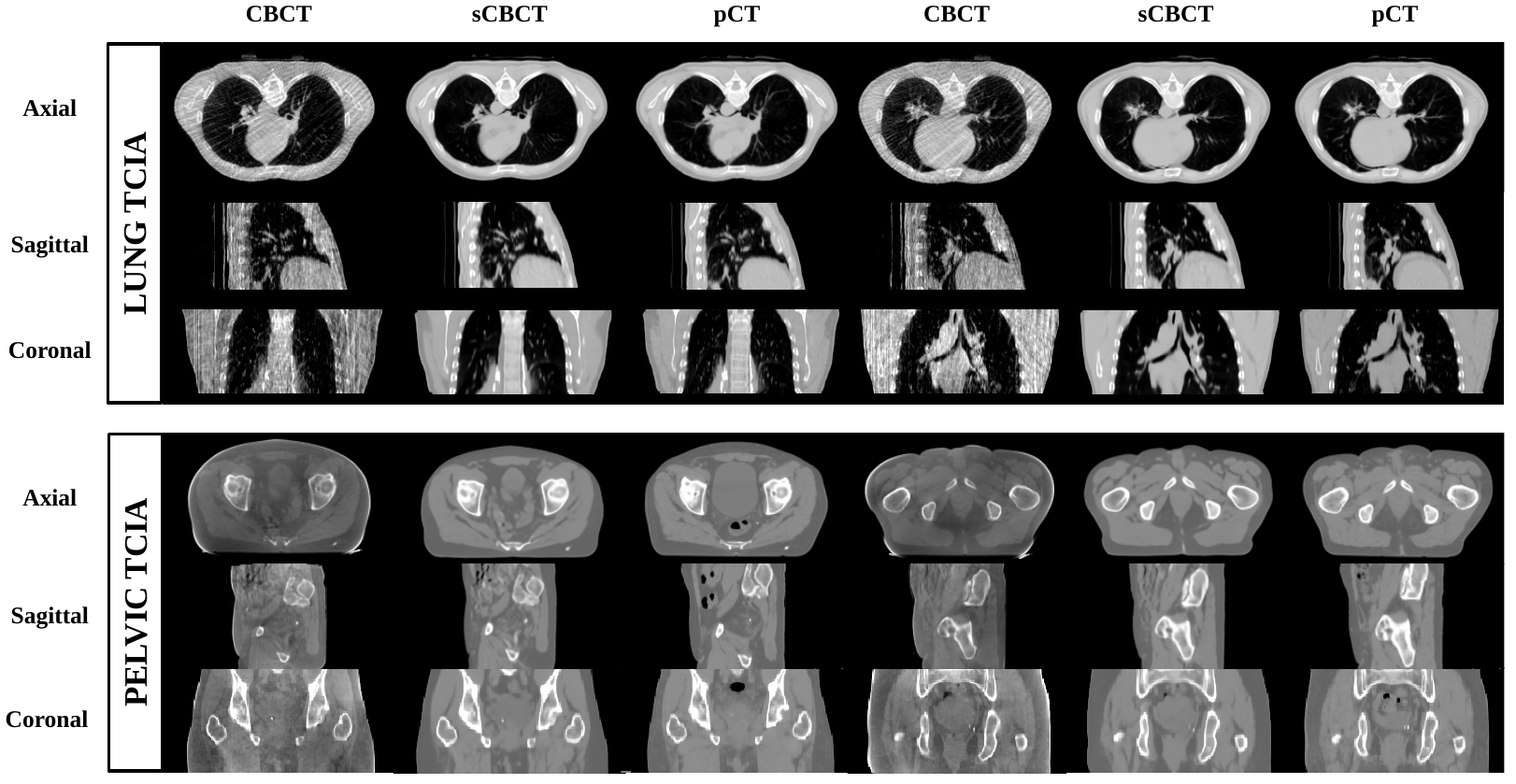}
    \caption{Representative volumetric results of RefineCBCT on the
    LUNG TCIA and PELVIC TCIA datasets. Two patient
    cases are shown for each dataset. For each case, axial, sagittal, and
    coronal views are presented for the original CBCT, the
    RefineCBCT-corrected image (sCBCT), and the corresponding planning CT
    (pCT). The sagittal and coronal views were reconstructed from the
    independently processed axial slices to qualitatively assess
    through-plane anatomical coherence.}
    \label{fig:mainresult}
\end{figure*}

Although CT and CBCT are inherently three-dimensional, RefineCBCT processes each axial slice independently using a 2D network. The corrected slices are subsequently reassembled into volumes to generate the sagittal and coronal views shown in Fig.~\ref{fig:mainresult}. Compared with the original CBCT images, the resulting sCBCT volumes show improved spatial uniformity, clearer anatomical structures, and fewer artifacts. These improvements are particularly evident on the \textit{LUNG TCIA} dataset, where streaking and shading artifacts are more pronounced. Minor differences between sCBCT and pCT may remain because of registration errors and inter-scan anatomical variation.

The slice-wise design reduces memory consumption and computational complexity relative to fully 3D processing, which is important for high-resolution CBCT refinement. Moreover, many CBCT artifacts are prominent within individual axial slices and can therefore be effectively addressed through in-plane correction. The reconstructed sagittal and coronal views in Fig.~\ref{fig:mainresult} do not show obvious discontinuities, suggesting that RefineCBCT maintains reasonable inter-slice coherence in the evaluated cases.

Nevertheless, the current framework does not explicitly model correlations between adjacent slices. Consequently, subtle through-plane inconsistencies may occur, particularly in regions with rapid anatomical changes, patient motion, or severe artifacts. The visual results alone cannot provide a quantitative assessment of volumetric consistency, and this limitation should be considered when applying the refined images to downstream 3D tasks.

Future work will extend RefineCBCT to 2.5D and 3D formulations. A 2.5D model could incorporate neighboring slices while retaining relatively low computational cost, whereas a fully 3D model could directly learn volumetric context and enforce through-plane structural continuity. Cross-slice consistency losses may provide another efficient alternative. These extensions will be evaluated using volumetric consistency measures and downstream tasks such as 3D segmentation, registration, contouring, and dose calculation.

\subsection{Effects of hyper-parameter values}

We further evaluated the sensitivity of RefineCBCT to the two loss weights, \(\lambda_{\mathrm{cyc}}\) and \(\lambda_{\mathrm{rec}}\), which control the cycle-consistency and reconstruction terms in Eqs.~(\ref{eq:loss-generation}) and (\ref{eq:loss-Gb}), respectively. Since both terms are essential to the joint generation--diffusion framework, cases with \(\lambda_{\mathrm{cyc}}=0\) or \(\lambda_{\mathrm{rec}}=0\) were not considered.

\begin{figure}[!htb]
\centering
\begin{subfigure}{0.9\textwidth}
    \centering
    \includegraphics[width=\textwidth]{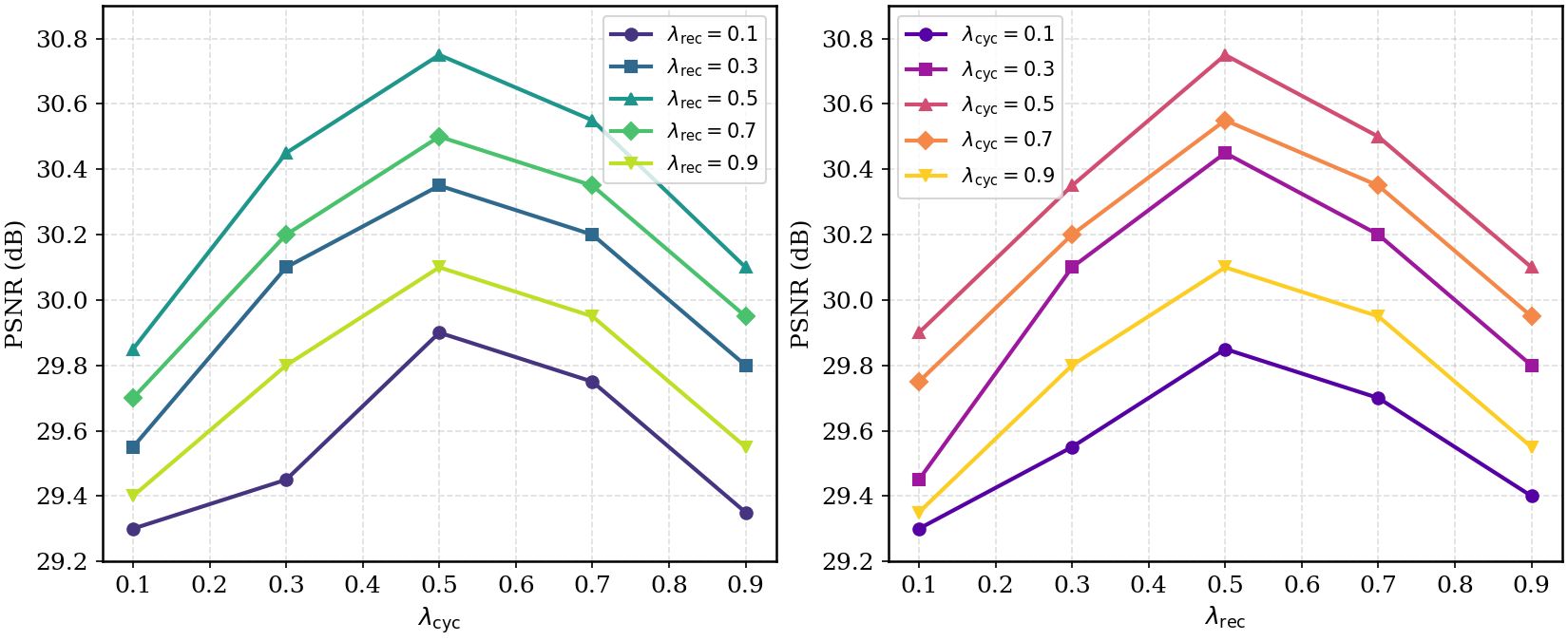}
    \caption{\small{LUNG TCIA}}
\end{subfigure}

\vspace{0.5em}

\begin{subfigure}{0.9\textwidth}
    \centering
    \includegraphics[width=\textwidth]{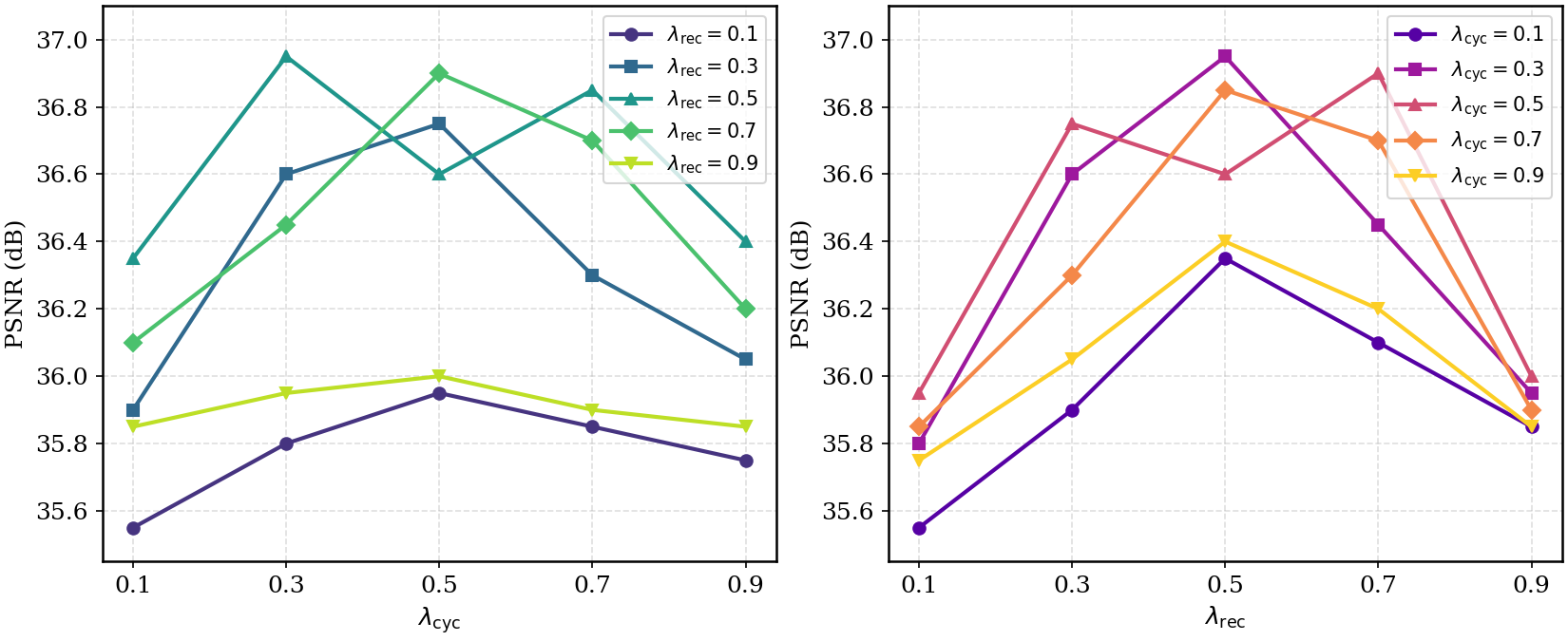}
    \caption{\small{{PELVIC TCIA}}}
\end{subfigure}

\caption{Hyper-parameter sensitivity analysis of RefineCBCT on two public datasets. PSNR is reported under different combinations of \(\lambda_{\mathrm{cyc}}\) and \(\lambda_{\mathrm{rec}}\).}
\label{fig:para_analysis}
\end{figure}

Figure~\ref{fig:para_analysis} shows the PSNR under different \((\lambda_{\mathrm{cyc}}, \lambda_{\mathrm{rec}})\) settings on the LUNG TCIA and \textit{PELVIC TCIA} datasets. On both datasets, the best performance is achieved with moderate and relatively balanced weight combinations rather than extreme values. Compared with \textit{PELVIC TCIA}, LUNG TCIA shows larger performance variations across different settings, indicating higher sensitivity to hyper-parameter selection. This may be due to the more complex anatomy and stronger artifacts in thoracic CBCT, which require a better balance between structural consistency and reconstruction fidelity.

These results suggest that RefineCBCT is reasonably robust to moderate changes in \(\lambda_{\mathrm{cyc}}\) and \(\lambda_{\mathrm{rec}}\), while still benefiting from dataset-specific tuning. Based on this analysis, we set \((\lambda_{\mathrm{cyc}}, \lambda_{\mathrm{rec}})=(0.5, 0.5)\) for LUNG TCIA and \((0.6, 0.4)\) for \textit{PELVIC TCIA} in the subsequent experiments.



\subsection{Implementation details} \label{ssec:details}

For each input CT slice \(c\), a guidance representation \(g = G_{\alpha}(c)\) was first generated and used to guide the subsequent refinement process. During diffusion sampling, the CBCT branch was initialized with Gaussian noise and iteratively updated for \(K\) denoising steps. At each step \(t\), the diffusion model \(G_{\beta}\) was conditioned on the concatenated input \([x_t, g]\), and the previous state \(x_{t-1}\) was sampled from the corresponding closed-form posterior distribution. Image synthesis was performed on a slice-by-slice basis, and the resulting slices were optionally reassembled into volumetric images for visualization and evaluation.

To balance reconstruction fidelity and computational efficiency, we adopted a short diffusion schedule. Unless otherwise specified, the number of sampling steps was set to \(K = 4\), which provided a favorable trade-off between image quality and inference time in our experiments. Exponential moving average (EMA)-updated generators were used during validation and testing to improve training stability and output consistency. The combination of a short diffusion process, adversarial optimization, and CT-guided pseudo-label supervision enabled efficient artifact reduction while preserving anatomically relevant structures.

We further investigated the sensitivity of the model to the diffusion length by evaluating \(K \in \{2, 4, 8\}\). Based on this analysis, \(K = 4\) was selected as the default configuration, as increasing the number of diffusion steps beyond this value led to higher computational cost with only marginal perceptual improvement.


\section{Discussion}
\label{sec:discussion}

In this study, we proposed RefineCBCT, an \textit{unpaired} CBCT refinement framework that combines pseudo-label guidance with short-step diffusion to improve image quality while preserving anatomical structures. Across two public datasets, RefineCBCT achieved better overall performance than several GAN-based and diffusion-based baseline methods. The method reduced common CBCT artifacts, improved visual image quality, and produced images with CT values that were closer to planning CT. These results suggest that RefineCBCT is a promising approach for CBCT enhancement in realistic unpaired training settings.

An \textit{important} feature of this work is that CBCT enhancement is treated as a refinement task rather than an unrestricted image generation. In medical imaging, good visual appearance alone is \textbf{not enough}; the enhanced image \textit{must} also preserve patient-specific anatomy. In RefineCBCT, the \textit{pseudo-label translation module} provides structural guidance for the diffusion process, and cycle-consistency helps reduce anatomical distortion during unpaired learning. The improvements in SSIM, visual quality, and HU consistency suggest that this design helps balance artifact reduction with anatomical fidelity.

The performance gain was \textit{more} obvious on the LUNG TCIA dataset than on the PELVIC TCIA dataset. This is likely because lung CBCT images usually contain more severe artifacts, making refinement more challenging but also allowing greater room for improvement. Still, the method also showed stable benefits on the pelvic dataset, which suggests that it can generalize across different anatomical sites. In addition, the HU analyses showed that the refined images were closer to planning CT in both bone and soft-tissue regions, which is encouraging for applications that require more reliable intensity information.

Another advantage of RefineCBCT is its \textbf{efficiency}. Standard diffusion models often require many sampling steps, which can \textit{limit} their use in clinical workflows. By using only a small number of diffusion steps, \textbf{our method reduces inference cost while still achieving strong image refinement performance}. The ablation study further showed that the pseudo-label generation module and cycle-consistency loss both made clear contributions to the final results, indicating that the full framework is \textit{necessary} to obtain the best balance between quality and efficiency.

\textbf{This study has several limitations}. i) the experiments were performed on two public datasets with a limited number of patients, and the model was trained in a 2D slice-by-slice manner rather than in full 3D;
ii) planning CT was used as the reference, but mismatch between CBCT and pCT may still affect evaluation;
iii) although image quality and HU consistency improved, we did not test downstream clinical tasks such as segmentation, dose calculation, or adaptive radiotherapy planning. Therefore, the present results should be interpreted as showing technical feasibility rather than direct clinical benefit. In addition, the robustness of the method across scanners, institutions, and imaging protocols still needs further validation.

In conclusion, RefineCBCT provides an \textbf{effective} and \textbf{efficient} method for \textit{unpaired} CBCT refinement. By combining pseudo-label guidance with short-step diffusion, it improves image quality, reduces artifacts, and better preserves anatomical structure without requiring paired CBCT-CT training data. These findings support the potential of clinically oriented diffusion-based refinement for CBCT enhancement and motivate future validation in larger and more diverse clinical settings.

\section{Methods}
\label{sec:methodology}

\subsection{Architecture Overview}
\label{ssec:overview}

\begin{figure*}[!htb]
  \centering
   \includegraphics[width=1.0\linewidth]{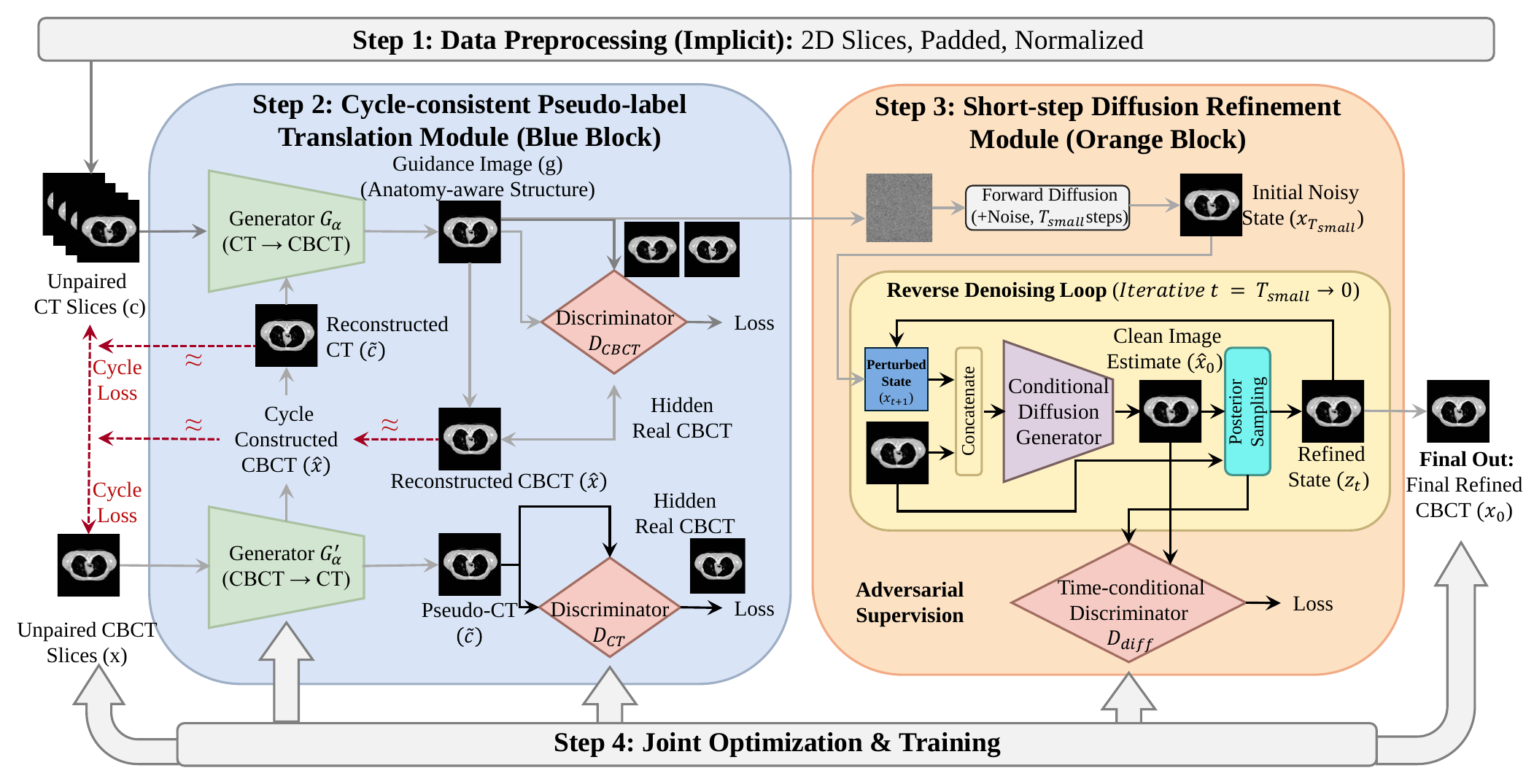}
   \caption{Technical details of RefineCBCT.}
   \label{fig:workflow}
\end{figure*}

We propose \textbf{RefineCBCT}, introduced a pseudo-label diffusion model for \emph{unpaired} cone-beam CT (CBCT) refinement task. The method is designed to improve CBCT image quality while preserving anatomical structures, with particular emphasis on \emph{robustness}, \emph{computational efficiency}, and practical potential for clinical use. As illustrated in Figure~\ref{fig:workflow}, RefineCBCT integrates two tightly coupled components:
i) a cycle-consistent pseudo-label translation module that bridges the CT and CBCT domains to provide anatomy-aware guidance, and
ii) a short-step conditional diffusion refinement module that progressively suppresses artifacts and restores CT-like image quality.

The pseudo-label translation module contains two image-to-image generators, CT$\rightarrow$CBCT and CBCT$\rightarrow$CT, each paired with a local PatchGAN-style discriminator. This branch produces pseudo-labels that provide domain-consistent structural guidance without requiring paired CBCT/CT supervision. The diffusion refinement module consists of a conditional generator operating with a small number of diffusion steps and a time-conditional discriminator that evaluates adjacent noisy states at a given timestep. To maintain a simple and efficient pipeline, all inputs are processed as single 2D slices, padded to a uniform spatial size and normalized to $[-1,1]$.

\textbf{Pseudo-label translation module.}
Given unpaired CT $c$ and CBCT $x$, the CT$\rightarrow$CBCT generator produces a guidance image $g$, while the CBCT$\rightarrow$CT generator produces a pseudo-CT image $\tilde c$. These translated images provide complementary supervision for bridging the appearance gap between the two domains. To reduce structural distortion and encourage anatomical consistency, cycle consistency is enforced by reconstructing
$\hat c = G'_{\alpha}(G_{\alpha}(c))$ and
$\hat x = G_{\alpha}(G'_{\alpha}(x))$.
In this way, the pseudo-label branch supplies anatomically meaningful guidance to the downstream diffusion refinement process.

\textbf{Short-step diffusion refinement module.}
Conditioned on the guidance image $g$, RefineCBCT refines the input CBCT through a small number of diffusion steps. At diffusion step $t$, a perturbed CBCT state $x_{t+1}$ is concatenated with $g$ and fed into the diffusion generator to predict the clean image estimate $\hat x_0$. A closed-form posterior is then used to obtain the refined sample $x_t$, enabling progressive restoration from noisy intermediate states to the final corrected image. Compared with conventional long-chain diffusion models, this short-step design substantially reduces inference cost while retaining the stability and strong denoising capability of diffusion-based generation, making the framework more suitable for time-constrained clinical scenarios.

\textbf{Adversarial learning and joint optimization.}
To further improve realism and structural fidelity, the diffusion branch is trained together with a time-conditional discriminator, which distinguishes real forward diffusion pairs $(x_t, x_{t+1})$ from model-derived posterior pairs generated from $\hat x_0$. Meanwhile, the pseudo-label translation module is supervised by local discriminators that encourage realistic cross-domain mapping. During training, each iteration alternates updates of the diffusion discriminator, the translation discriminators, and then all generators using adversarial, cycle-consistency, and reconstruction objectives. In addition, exponential moving average (EMA) is applied to the generators to stabilize optimization and improve sampling quality.

\subsection{Pseudo-label Translation Module}
\label{ssec:Generation}

A key challenge in unpaired CBCT refinement is the lack of voxel-wise aligned CBCT/CT supervision. To address this, RefineCBCT incorporates a \emph{pseudo-label translation module} that learns bidirectional mapping between the CBCT domain $\mathcal{X}$ and the CT domain $\mathcal{C}$. This module is designed to provide anatomy-aware cross-domain guidance for the downstream diffusion refinement process while preserving structural consistency under unpaired training. By translating between CT and CBCT in both directions, the module reduces the domain gap and supplies pseudo-labels that stabilize subsequent CBCT refinement.

Given a CT slice $c \sim \mathcal{C}$ and a CBCT slice $x \sim \mathcal{X}$, the CT$\!\to\!$CBCT generator $G_{\alpha}$ produces a guidance image $g$, and the CBCT$\!\to\!$CT generator $G'_{\alpha}$ produces a pseudo-CT image $\tilde c$:
\begin{equation}
g = G_{\alpha}(c), \qquad \tilde c = G'_{\alpha}(x).
\label{eq:guidance-and-pseudoct}
\end{equation}
Here, $g$ is a CT-derived image translated into the CBCT domain and serves as structural guidance for the diffusion branch, while $\tilde c$ is a CT-like pseudo-label estimated from the input CBCT. These two outputs provide complementary cross-domain information that helps RefineCBCT suppress artifacts while maintaining anatomical fidelity.

To ensure that translation preserves image content rather than merely matching target-domain appearance, we impose cycle consistency in both directions. Specifically, a CT image is translated to the CBCT domain and then mapped back to CT, while a CBCT image is translated to the CT domain and then mapped back to CBCT:
\begin{equation}
\hat c = G'_{\alpha}(G_{\alpha}(c)), \qquad \hat x = G_{\alpha}(G'_{\alpha}(x)).
\label{eq:cycle-recon}
\end{equation}
The reconstructed images $\hat c$ and $\hat x$ are constrained to remain close to the original inputs. This cycle-consistent design encourages near-invertible translation, preserves anatomical geometry across domains, and reduces the risk of hallucinated structures, which is critical for clinically meaningful CBCT refinement.

To improve domain realism, the pseudo-label translation module is trained adversarially in both directions. A CBCT discriminator $D_{\alpha}$ distinguishes real CBCT slices from translated samples $G_{\alpha}(c)$, while a CT discriminator $D'_{\alpha}$ distinguishes real CT slices from translated samples $G'_{\alpha}(x)$. Using the non-saturating GAN objective with a softplus surrogate, the CBCT-side discriminator loss is defined as
\begin{equation}
\begin{aligned}
\mathcal{L}_{D_{\alpha}}
&= \mathbb{E}_{x\sim\mathcal{X}}\!\big[\operatorname{softplus}(D_{\alpha}(x))\big] \\
&\quad + \mathbb{E}_{c\sim\mathcal{C}}\!\big[\operatorname{softplus}(-D_{\alpha}(G_{\alpha}(c)))\big],
\end{aligned}
\label{eq:loss-Da}
\end{equation}
where $\operatorname{softplus}(u)=\log(1+e^u)$ and $\mathbb{E}[\cdot]$ denotes expectation over the corresponding data distribution. This objective encourages translated samples $G_{\alpha}(c)$ to match the appearance distribution of real CBCT images.

The corresponding CT$\!\to\!$CBCT generator loss is
\begin{equation}
\mathcal{L}_{G_{\alpha}}
= \mathbb{E}_{c\sim\mathcal{C}}\!\big[\operatorname{softplus}(-D_{\alpha}(G_{\alpha}(c)))\big],
\label{eq:loss-Ga}
\end{equation}
which encourages the translated guidance image $g$ to be indistinguishable from real CBCT data while preserving anatomically relevant structures.

Similarly, the CT-side discriminator loss is
\begin{equation}
\begin{aligned}
\mathcal{L}_{D'_{\alpha}}
&= \mathbb{E}_{c\sim\mathcal{C}}\!\big[\operatorname{softplus}(D'_{\alpha}(c))\big] \\
&\quad + \mathbb{E}_{x\sim\mathcal{X}}\!\big[\operatorname{softplus}(-D'_{\alpha}(G'_{\alpha}(x)))\big],
\end{aligned}
\label{eq:loss-Dpa}
\end{equation}
and the CBCT$\!\to\!$CT generator is optimized with
\begin{equation}
\mathcal{L}_{G'_{\alpha}}
= \mathbb{E}_{x\sim\mathcal{X}}\!\big[\operatorname{softplus}(-D'_{\alpha}(G'_{\alpha}(x)))\big].
\label{eq:loss-Gpa}
\end{equation}
These adversarial objectives improve translation fidelity in both domains and help produce more realistic and stable pseudo-labels for guidance.

To preserve anatomical consistency across domains, we further use the cycle-consistency loss
\begin{equation}
\mathcal{L}_{\mathrm{cyc}} =
\big\|G'_{\alpha}(G_{\alpha}(c)) - c\big\|_1
+
\big\|G_{\alpha}(G'_{\alpha}(x)) - x\big\|_1,
\label{eq:loss-cyc}
\end{equation}
where $\|\cdot\|_1$ denotes the elementwise $L_1$ norm. The first term penalizes deviations in the CT$\!\to\!$CBCT$\!\to\!$CT cycle, and the second penalizes deviations in the CBCT$\!\to\!$CT$\!\to\!$CBCT cycle. This constraint is essential for maintaining structural fidelity and reducing content drift that adversarial learning alone may not prevent.

The pseudo-label translation module is optimized by combining the adversarial objectives in both directions with the cycle-consistency constraint:
\begin{equation}
\mathcal{L}_{\mathrm{Generation}} =
\mathcal{L}_{G_{\alpha}} + \mathcal{L}_{D_{\alpha}}
+ \mathcal{L}_{G'_{\alpha}} + \mathcal{L}_{D'_{\alpha}}
+ \lambda_{\mathrm{cyc}}\,\mathcal{L}_{\mathrm{cyc}},
\label{eq:loss-generation}
\end{equation}
where $\lambda_{\mathrm{cyc}}>0$ controls the strength of the cycle-consistency term. Through this bidirectional translation process, RefineCBCT learns anatomically meaningful pseudo-labels that bridge the CT and CBCT domains and provide robust structural guidance for the subsequent diffusion refinement module.

.

\subsection{Short-step Diffusion Refinement Module}
\label{ssec:Diffusive}

Building on the anatomy-aware guidance produced by the pseudo-label translation module, RefineCBCT adopts a \emph{short-step diffusion refinement module} to progressively suppress CBCT artifacts and recover CT-like image quality. Unlike conventional diffusion models that rely on long denoising trajectories, our design uses a truncated variance-preserving process with only $K$ reverse steps, which substantially reduces computational cost while retaining stable iterative refinement. This short-step strategy is particularly suitable for clinical CBCT correction, where both robustness and inference efficiency are important.

Let $\{\beta_t\}_{t=1}^{K}$ denote the variance schedule, with $\alpha_t = 1-\beta_t$ and $\bar{\alpha}_t = \prod_{s=1}^{t}\alpha_s$. For an input CBCT slice $x_0$, the forward diffusion process gradually perturbs the image with Gaussian noise to obtain a noisy sample $x_t$ at step $t$:
\begin{equation}
x_t = \sqrt{\bar\alpha_t}\,x_0 + \sqrt{1-\bar\alpha_t}\,\varepsilon,
\qquad \varepsilon\sim\mathcal{N}(0,I).
\label{eq:forward-diffusion}
\end{equation}
Here, $x_0$ denotes the input CBCT slice, $x_t$ is its noisy version at diffusion step $t$, and $\varepsilon$ is standard Gaussian noise. The coefficient $\bar\alpha_t$ controls the signal-to-noise ratio during corruption. These noisy states define the forward diffusion trajectory used to train the reverse refinement process.

At reverse step $t+1 \rightarrow t$, the diffusion generator $G_{\beta}$ predicts a refined estimate of the clean image conditioned on the current noisy sample $x_{t+1}$ and the guidance image $g$ produced by the pseudo-label translation module:
\begin{equation}
\hat{x}_0 = G_{\beta}\big([x_{t+1},\, g],\, t,\, z\big),
\qquad z\sim\mathcal{N}(0,I),
\label{eq:generator-pred}
\end{equation}
where $[x_{t+1}, g]$ denotes channel-wise concatenation, $t$ is the timestep embedding, $z$ is latent Gaussian noise, and $\hat{x}_0$ is the predicted refined image. Conditioning on $g$ narrows the solution space of reverse diffusion by injecting anatomy-preserving structural guidance, thereby reducing ambiguity and improving refinement stability.

Given the predicted image $\hat{x}_0$, the previous latent state is sampled from the corresponding posterior distribution:
\begin{equation}
x_t \sim q(x_t \mid x_{t+1}, \hat{x}_0)
= \mathcal{N}\!\big(\mu_t(\hat{x}_0, x_{t+1}),\, \sigma_t^2 I\big),
\label{eq:posterior}
\end{equation}
where $\mu_t(\hat{x}_0, x_{t+1})$ and $\sigma_t^2$ are determined by the variance-preserving diffusion schedule. This posterior sampling step enables progressive refinement along a short reverse trajectory, allowing RefineCBCT to improve image quality with fewer denoising iterations.

To enhance temporal realism of the reverse dynamics, we introduce a diffusion discriminator $D_{\beta}$ that distinguishes true forward-diffusion transitions from model-generated reverse transitions. Specifically, $(x_t, x_{t+1})$ forms a real diffusion pair from the forward process, whereas $\tilde{x}_t \sim q(x_t \mid x_{t+1}, \hat{x}_0)$ is sampled from the model posterior. The discriminator is trained with an adversarial objective regularized by an $R_1$ gradient penalty:
\begin{equation}
\begin{aligned}
\mathcal{L}_{D_{\beta}}
&= \mathbb{E}\!\big[\operatorname{softplus}(-D_{\beta}(x_t, t, x_{t+1}))\big] \\
&\quad + \mathbb{E}\!\big[\operatorname{softplus}(D_{\beta}(\tilde{x}_t, t, x_{t+1}))\big] \\
&\quad + \frac{\gamma}{2}\,\mathbb{E}\!\big[\|\nabla_{x_t} D_{\beta}(x_t, t, x_{t+1})\|_2^2\big],
\end{aligned}
\label{eq:loss-Db}
\end{equation}
where $\gamma > 0$ is the coefficient of the gradient penalty. This objective encourages the generated reverse transitions to match the local dynamics of the true diffusion process while stabilizing discriminator training.

The generator is optimized to fool the discriminator and reconstruct the underlying CBCT content:
\begin{equation}
\begin{aligned}
\mathcal{L}_{G_{\beta}}
&= \mathbb{E}\!\big[\operatorname{softplus}(-D_{\beta}(\tilde{x}_t, t, x_{t+1}))\big] \\
&\quad + \lambda_{\mathrm{rec}}\,\|\hat{x}_0 - x_0\|_1,
\end{aligned}
\label{eq:loss-Gb}
\end{equation}
where $\lambda_{\mathrm{rec}} > 0$ controls the reconstruction term. The adversarial term encourages realistic reverse-step transitions, while the $L_1$ reconstruction term preserves image content and prevents the refinement process from over-relying on guidance alone.

Overall, the short-step diffusion refinement module complements the pseudo-label translation module in a tightly coupled manner. The translated guidance image $g$ provides anatomically meaningful structural constraints that narrow the reverse diffusion search space, while the short-step refinement process progressively removes artifacts with limited computational overhead. This design enables RefineCBCT to achieve robust CBCT enhancement under unpaired supervision while remaining efficient enough for practical clinical deployment.

\subsection{Total Loss and Training Details}
\label{ssec:loss}

To jointly optimize anatomy-preserving pseudo-label estimation and efficient CBCT refinement, RefineCBCT combines the objectives of the pseudo-label translation module and the short-step diffusion refinement module into a unified training framework. The overall objective is defined as
\begin{equation}
\mathcal{L}_{\mathrm{RefineCBCT}}
=
\mathcal{L}_{\mathrm{trans}}
+
\mathcal{L}_{G_{\beta}}
+
\mathcal{L}_{D_{\beta}},
\label{eq:loss-total}
\end{equation}
where $\mathcal{L}_{\mathrm{trans}}$ denotes the total loss of the pseudo-label translation module, including bidirectional adversarial translation and cycle-consistency regularization, while $\mathcal{L}_{G_{\beta}}$ and $\mathcal{L}_{D_{\beta}}$ are the generator and discriminator objectives of the short-step diffusion refinement module, respectively.

Specifically, $\mathcal{L}_{\mathrm{trans}}$ encourages realistic CT$\!\leftrightarrow\!$CBCT translation while preserving anatomical structure across domains, enabling the model to generate the guidance image $g$ and pseudo-CT image $\tilde{c}$ from unpaired training data. The diffusion generator objective $\mathcal{L}_{G_{\beta}}$ promotes anatomically guided reverse-step refinement and preserves image fidelity through the reconstruction term, whereas the discriminator objective $\mathcal{L}_{D_{\beta}}$ enforces temporal realism of the diffusion transitions under a non-saturating adversarial loss with $R_1$ regularization. Together, these components couple pseudo-label guidance with efficient denoising, allowing RefineCBCT to suppress CBCT artifacts while maintaining anatomically consistent structures and reliable intensity recovery.

During training, the translation generators $(G_{\alpha}, G'_{\alpha})$, diffusion generator $G_{\beta}$, and discriminators $(D_{\alpha}, D'_{\alpha}, D_{\beta})$ are optimized in an alternating manner. For each unpaired minibatch, the pseudo-label translation module first produces the guidance image $g$ and pseudo-CT $\tilde{c}$, which are then used to condition the short-step diffusion refinement process. In this way, the pseudo-labels are updated jointly with the diffusion model rather than being pre-computed offline, allowing the two modules to co-adapt throughout training. To improve optimization stability and inference robustness, we maintain an exponential moving average (EMA) of the generator parameters and use the EMA model for validation and inference.

The balance between anatomical preservation and refinement fidelity is controlled by the loss weights in each module. In particular, $\lambda_{\mathrm{cyc}}$ in the translation objective regulates the strength of structural consistency across CT and CBCT domains, while $\lambda_{\mathrm{rec}}$ in the diffusion generator objective balances guidance-driven refinement against intensity-faithful reconstruction. This joint optimization design enables RefineCBCT to learn structure-consistent pseudo-labels and efficient denoising trajectories simultaneously, supporting robust and practical CBCT refinement for clinical use.


The full training routine is summarized in Algorithm~\ref{alg:RefineCBCT}.
\begin{algorithm}[H]
\caption{Training procedure of RefineCBCT}
\label{alg:RefineCBCT}
\begin{algorithmic}
\State \textbf{Input}: Unpaired minibatches $x \sim \mathcal{X}$ and $c \sim \mathcal{C}$; diffusion steps $K$; EMA decay $\tau$.
\State \textbf{Output}: Updated parameters of $G_{\alpha}, G'_{\alpha}, G_{\beta}, D_{\alpha}, D'_{\alpha}, D_{\beta}$.

\State \textbf{1: Pseudo-label translation}
\State \hspace{0.5cm} Generate guidance image $g$ and pseudo-CT $\tilde{c}$ using Eq.~(\ref{eq:guidance-and-pseudoct}).
\State \hspace{0.5cm} Compute cycle reconstructions $\hat{c}$ and $\hat{x}$ using Eq.~(\ref{eq:cycle-recon}).

\State \textbf{2: Forward diffusion pair construction}
\State \hspace{0.5cm} Sample timestep $t \sim \{0,\ldots,K-1\}$.
\State \hspace{0.5cm} Construct adjacent noisy states $(x_t, x_{t+1})$ according to Eq.~(\ref{eq:forward-diffusion}).

\State \textbf{3: Guidance-conditioned reverse prediction}
\State \hspace{0.5cm} Predict refined image estimate $\hat{x}_0$ from $[x_{t+1}, g]$ using Eq.~(\ref{eq:generator-pred}).
\State \hspace{0.5cm} Sample posterior state $\tilde{x}_t$ using Eq.~(\ref{eq:posterior}).

\State \textbf{4: Discriminator update}
\State \hspace{0.5cm} Update $D_{\alpha}$, $D'_{\alpha}$, and $D_{\beta}$ using Eqs.~(\ref{eq:loss-Da}), (\ref{eq:loss-Dpa}), and (\ref{eq:loss-Db}).

\State \textbf{5: Generator update}
\State \hspace{0.5cm} Update $G_{\alpha}$, $G'_{\alpha}$, and $G_{\beta}$ using Eqs.~(\ref{eq:loss-Ga}), (\ref{eq:loss-Gpa}), and (\ref{eq:loss-Gb}).
\State \hspace{0.5cm} Apply cycle-consistency regularization using Eq.~(\ref{eq:loss-cyc}).

\State \textbf{6: Stabilization}
\State \hspace{0.5cm} Compute the total objective in Eq.~(\ref{eq:loss-total}).
\State \hspace{0.5cm} Update the exponential moving average of generator parameters with decay $\tau$.
\end{algorithmic}
\end{algorithm}


\section{Declarations}\label{sec:Declarations}

\noindent \textbf{Data Availability}

This work did not generate any new physical materials.The trained model weights and configuration fles (for training and inference) have been published. The dataset can be download from \url{https://www.cancerimagingarchive.net/}. Additional artifacts can be provided by the first author upon reasonable request.

\noindent \textbf{Code availability}

Code has publicly available at: \url{https://github.com/Lai-KK/RefineCBCT.git}.

\noindent \textbf{Ethics approval and consent to participate}

This study received approval from the Institutional Review Board (IRB) of Xiangyang Central Hospital,China,under protocol number.

\noindent \textbf{Author contribution}

QI LAI: Conceptualization, Methodology, Formal analysis, Writing-Original draft; 
YUTONG HE: Methodology, Validation, Experiments;
DONG LIANG: Data curation, Investigation, Resources;
LEI WANG: Writing Review,Funding acquisition;
YAN YAN: Software,Validation,Data curation;
YANZHONG MA: Investigation, Data curation;
YIMIN ZHOU: Investigation, Resources, Supervision;
ZHICHENG LI: Data curation, Investigation, Resources;
XIAOKUN LIANG: Writing Review, Funding acquisition;
LEI JIN: Writing Review, Funding acquisition.

\noindent \textbf{Conflict of interest/Competing interests}

The authors declare that they have no conflicts of interest or competing interests related to this work.

\noindent \textbf{Acknowledgments}

This work was supported by the National Key Research and Development Program of China (2024YFF1206901); in part by the Shenzhen General Program under Grant Ref. JCYJ20240813154703005; in part by the National Natural Science Fund for Young Scientists Fund Program C (82502447).


\end{document}